\documentclass[10pt,twocolumn,letterpaper]{article}

\usepackage[pagenumbers]{wacv} 

\usepackage{graphicx}
\usepackage{amsmath}
\usepackage{amssymb}
\usepackage{booktabs}
\usepackage{mathtools }

\usepackage[pagebackref,breaklinks,colorlinks]{hyperref}

\usepackage[capitalize]{cleveref}
\crefname{section}{Sec.}{Secs.}
\Crefname{section}{Section}{Sections}
\Crefname{table}{Table}{Tables}
\crefname{table}{Tab.}{Tabs.}

\DeclareMathOperator*{\argmax}{arg\,max} 
\usepackage{bbm}
\usepackage[ruled,vlined,linesnumbered]{algorithm2e}
\DontPrintSemicolon
\SetKwInput{Input}{Input}
\SetKwInput{Output}{Output}
\usepackage{wrapfig}
\usepackage{multirow}
\usepackage{bm}

\usepackage{color}

\newcommand{\blue}[1]{{\color{black}{#1}}}

\def\wacvPaperID{664} 
\def\confName{WACV}
\def\confYear{2025}

\begin{document}

\title{Federated Source-free Domain Adaptation for Classification:\\Weighted Cluster Aggregation for Unlabeled Data}

\author{Junki Mori\\
NEC Corporation, Kyoto University\\
Kawasaki, Japan\\
{\tt\small junki.mori@nec.com}
\and
Kosuke Kihara \\
NEC Corporation \\
Kawasaki, Japan\\
{\tt\small kosuke-kihara@nec.com}
\and
Taiki Miyagawa \\
NEC Corporation \\ 
Kawasaki, Japan\\
{\tt\small miyagawataik@nec.com}
\and
Akinori F. Ebihara \\
NEC Corporation \\
Kawasaki, Japan\\
{\tt\small aebihara@nec.com}
\and
Isamu Teranishi \\
NEC Corporation \\
Kawasaki, Japan\\
{\tt\small teranisi@nec.com}
\and
Hisashi Kashima \\
Kyoto University \\
Kyoto, Japan\\
{\tt\small kashima@i.kyoto-u.ac.jp}
}
\maketitle

\begin{abstract}

\textit{Federated learning} (FL) commonly assumes that the server or some clients have labeled data, which is often impractical due to annotation costs and privacy concerns. Addressing this problem, we focus on a source-free domain adaptation task, where (1) the server holds a pre-trained model on labeled source domain data, (2) clients possess only unlabeled data from various target domains, and (3) the server and clients cannot access the source data in the adaptation phase. This task is known as \textit{Federated source-Free Domain Adaptation} (FFREEDA). 
Specifically, we focus on classification tasks, while the previous work solely studies semantic segmentation. 
Our contribution is the novel \textit{Federated learning with Weighted Cluster Aggregation} (FedWCA) method, designed to mitigate both domain shifts and privacy concerns with only unlabeled data. 
FedWCA comprises three phases: private and parameter-free clustering of clients to obtain domain-specific global models on the server, weighted aggregation of the global models for the clustered clients, and local domain adaptation with pseudo-labeling.
Experimental results show that FedWCA surpasses several existing methods and baselines in FFREEDA, establishing its effectiveness and practicality.

\end{abstract}

\section{Introduction}
\label{sec:intro}
Most existing studies on federated learning (FL), a privacy-preserving machine learning paradigm for distributed data, assume that labeled data is available at each client. 
However, this may not be the case for vision tasks which often necessitate cumbersome manual data annotation requiring domain expertise.
For instance, in applying FL to multi-location surveillance cameras, it is burdensome for administrators to manually annotate each image. While a server, typically hosted by organizations, is more likely to access labeled data, 
the included biometric information such as faces, makes direct use of this data in FL problematic due to privacy concerns.
The situation also exemplifies the challenges of differing domain characteristics, like brightness and background, based on the camera's location.

\begin{figure}[t]
    \centering
    \includegraphics[width=0.95\linewidth]{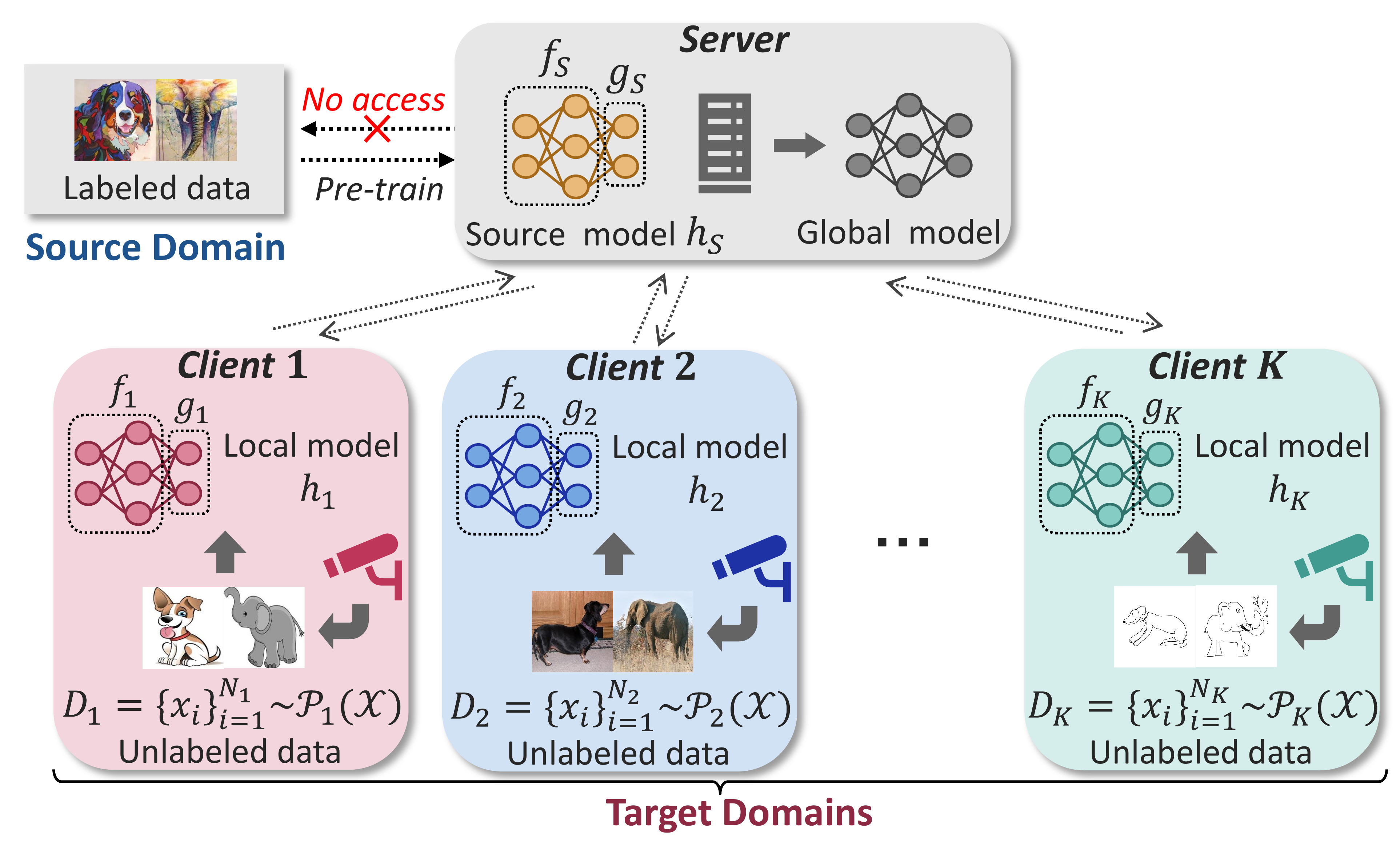}
    \caption{\textbf{Overview of our problem setting: FFREEDA \cite{shenaj_wacv2023_ladd}}. 
    A pre-trained model on a labeled \textit{source} domain dataset exists on the server, which becomes inaccessible following pre-training (\textit{source-free}).
    Clients possess unlabeled datasets from specific \textit{target} domains, which may or may not be identical.
    We aim to train personalized models tailored to each client's domain.
    }
    \label{fig:prob_set}
\end{figure}

Building on the preceding example, this study explores the practical yet less-investigated FL setting of \textit{Federated source-Free Domain Adaptation} (FFREEDA) \cite{shenaj_wacv2023_ladd}, specifically for classification tasks (\cref{fig:prob_set}). In FFREEDA, a server possesses a model pre-trained on labeled data from a source domain, but this data is not usable in later FL stages due to privacy issues or immediate deletion after pre-training. 
In addition, each client holds limited unlabeled data in a target domain, which 
may greatly differ from the source domain and the domains of other clients, thus creating an unsupervised multi-target setting.
FFREEDA aims for each client to train a classification model tailored to its domain, leveraging the source model and the knowledge from other clients. 

\begin{table*}[t]
    \caption{\textbf{Related problem settings.} 
    Our study focuses on classification tasks, tackling three distinct problem types: "No Labels," "Domain Shift," and "Personalization."
    $\triangle$ denotes specific characteristics depending on FL paradigms: while most Personalized FL research concentrates on label shifts, certain studies like \cite{partialfed,fedpcl} also tackle domain shifts. Semi-Supervised FL literature typically involves unlabeled data, presuming the presence of labeled data either on the server or within a subset of clients, a scenario also observed in some FDA studies.}
    \centering
    \small
    \begin{tabular}{lcccc}\hline
        \textbf{FL paradigm}
        & \textbf{No labels} & \ \textbf{Domain shift} \ & \ \textbf{Personalization} \ & \textbf{Task} \\ 
        \hline
        Personalized FL \cite{three,perfedavg,fedrep,pfedmb,fedfomo,fedamp,partialfed,fedpcl} & & $\triangle$ & \checkmark & Classification \\
        Semi-supervised FL \cite{semifl,cbafed,crl,fedmatch,fedssl,rscfed} & $\triangle$ & & & Classification \\
        Unsupervised FL \cite{furl,orchestra,protofl,som,uifca} & \checkmark & & & Representation learning/Clustering \\
        \multirow{2}{*}{FDA}  \cite{fada,fade,efficientfda,fedka,kd3a} & $\triangle$ & \checkmark & & Classification \\
        \hspace{18pt}  \cite{dualadapt,fruda} & $\triangle$ & \checkmark & \checkmark & Classification \\
        \multirow{2}{*}{FFREEDA} \hspace{2pt} LADD \cite{shenaj_wacv2023_ladd} & \checkmark & \checkmark & \checkmark & Semantic segmentation \\
        \hspace{45pt} \textbf{Ours} & \checkmark & \checkmark & \checkmark & Classification \\
        \hline
    \end{tabular}
    \label{table:related_works}
\end{table*}


We tackle two main challenges of FFREEDA: (1) fully unsupervised FL after pre-training and (2) domain shifts among clients. 
We propose a novel FFREEDA method for classification, \textit{Federated Learning with Weighted Cluster Aggregation} (FedWCA). 
This overcomes issues in the prior FFREEDA method \cite{shenaj_wacv2023_ladd} for semantic segmentation, including privacy issues connected to image style feature sharing during client clustering and limited performance (\cref{sec:experiment}) due to a cluster-restricted model aggregation and a simple pseudo-labeling.
Specifically, FedWCA comprises three key components. 
Firstly, to avert performance degradation from domain shift (2), we cluster clients based on the initial layer parameters of each client's locally adapted feature extractor, which contains domain information. 
This process relies only on model parameters, eliminating the need for additional information sharing and hyperparameters.
To further boost performance, we introduce the Weighted Cluster Aggregation (WCA) module, enabling clients to combine models from each cluster (domain) using personal weights. 
A novel metric is defined for computing cluster weights based solely on unlabeled data, considering the benefits for both individual clients and the overall group.
Finally, handling unlabeled target data (1) is achieved through prototype-based pseudo-labeling \cite{liang_icml2020_shot}. We improve it to retain global knowledge and produce more reliable pseudo-labels.
Empirical validation confirms the efficacy of these components, with the 
complete method outperforming all baselines, as detailed in \cref{sec:experiment}.

This work primarily contributes as follows:
\begin{itemize}
    \item We introduce a novel FFREEDA method for classification, named FedWCA, with three main components: (1) a private, parameter-free client clustering module, (2) a weighted cluster aggregation module for domain knowledge integration, and (3) a local adaptation module using an improved pseudo-labeling strategy.
    \item Empirical results show that FedWCA significantly enhances performance over all baselines, including the prior method \cite{shenaj_wacv2023_ladd}, across three multi-domain benchmark datasets: Digit-Five, PACS, and Office-Home.
\end{itemize}

\section{Related Work}
\label{sec:related_works}

\subsection{Source-free Domain Adaptation}
This paper extends \textit{Source-free domain adaptation} (SFDA), a type of unsupervised domain adaptation approach, to the FL setting.
SFDA involves adapting a pre-trained source model to an unlabeled target domain without accessing the source data. A fundamental SFDA method is pseudo-labeling for unlabeled target data \cite{liang_icml2020_shot,dipe,sfda-de,bmd,pmlr-v162-lee22c,Litrico_2023_CVPR}. 
A prototype-based pseudo-labelling method introduced by \textit{Source HypOthesis Transfer} (SHOT) \cite{liang_icml2020_shot}, is commonly employed to derive pseudo-labels, minimizing label noise induced by domain shifts \cite{dipe,sfda-de,bmd}.

\subsection{Federated Domain Adaptation}
FADA \cite{fada} is the pioneering work addressing \textit{Federated Domain Adaptation} (FDA) for a single target domain. 
Subsequent studies \cite{fade,efficientfda,fedka,kd3a} have also focused on a multi-source-single-target setting. 
DualAdapt \cite{dualadapt} explores a single-source-multi-target scenario, similar to this paper, where each client gets one target domain and the server holds a source domain. 
FRuDA \cite{fruda} considers sequentially adding new target domains to the federated system. 
Unlike our source-free setting, these studies typically assume access to source data during FL (\cref{table:related_works}).

\subsection{Federated Source-free Domain Adaptation}
Our model is inspired by \textit{Learning Across Domains and Devices} (LADD) \cite{shenaj_wacv2023_ladd}, the sole study addressing FFREEDA, which focuses on semantic segmentation.
In contrast, our model is designed for classification tasks, decreases the number of hyperparameters, alleviates privacy concerns by eliminating the need to transmit image style features from clients to servers for client clustering (\cref{sec:client_clustering}), integrates all domain information with client-specific weights (Sec.~\ref{aggregation_method}), generates more reliable pseudo-labels (\cref{sec:local_adaptation_module}), and thus improves accuracy (Sec.~\ref{sec:experiment}). 


\subsection{Federated Learning with Non-i.i.d. Data}
A major challenge in FL is data heterogeneity, namely the non-i.i.d. (independent and identically distributed) data distribution problem \cite{fedprox}. \textit{Personalized FL} \cite{PFL} addresses this by training models tailored to each client's unique data distribution \cite{three,perfedavg,fedrep,partialfed,pfedmb,fedfomo,fedamp}. 
\textit{Clustered FL}, a variant of personalized FL, also tackles non-i.i.d. data by clustering clients based on data distributions and aggregating models within clusters. However, existing clustered FL methods often require setting hyperparameters such as cluster numbers \cite{cfl,ifca,fedsim,fedgc,fedfmc}, thresholds \cite{fl+hc,fedseq}, iteration counts \cite{wscc}, and neighbor candidate list sizes \cite{panm}. 
Although these studies focus on non-i.i.d. data, they 
largely assume labeled data and emphasize label shifts more than 
domain shifts (\cref{table:related_works}).

\subsection{Federated Learning with Unlabeled Data}
In the realm of FL with unlabeled data, such as \textit{semi-supervised FL}, two main settings exist: one where the server has labeled data and clients only have unlabeled data \cite{semifl,cbafed,crl,fedmatch} and another where some clients possess fully labeled data while others have entirely unlabeled data \cite{fedssl,rscfed,semifedseg}. 
Our work explores a fully unsupervised setting for both server and client sides after the source pre-training phase. 
\textit{Unsupervised FL}, which treats fully unsupervised settings throughout training, typically focuses on representation learning \cite{furl,orchestra,protofl} or clustering tasks \cite{som,uifca}, unlike our emphasis on classification tasks (\cref{table:related_works}).

\section{Problem Formulation}%
\label{sec:problem_formulation}

Our FFREEDA is illustrated in \cref{fig:prob_set}. In our FFREEDA, a server and \blue{$K$ clients (indexed by $k$)} address an $M$-way classification problem with a feature space 
$\mathcal{X}$
and a label space $\mathcal{Y}$. 
Each client \blue{$k$} holds an unlabeled dataset \blue{$D_k=\{x_{i}\}_{i=1}^{N_k} \sim \mathcal{P}_k(\mathcal{X})$} from multiple target domains (client side in \cref{fig:prob_set}), where \blue{$\mathcal{P}_k$} represent the respective data distributions. 
Distinct domain shifts exist among the clients, denoted by
\blue{$\mathcal{P}_k \neq \mathcal{P}_l$}. 
Note that certain clients may share the same domain.


The server has a source model $h_S\colon \mathcal{X} \rightarrow \mathbb{R}^{M}$ pre-trained on a labeled dataset from the source domain whose data distribution differs from \blue{$\mathcal{P}_k$} for any \blue{$k$} (server side in \cref{fig:prob_set}). The labeled source dataset is discarded or retained privately, preventing further access.  
The model $h_S$ has two modules: $h_S = g_S \circ f_S$, where $f_S \colon \mathcal{X} \rightarrow \mathbb{R}^q$ is a feature extractor, $g_S \colon \mathbb{R}^q \rightarrow \mathbb{R}^{M}$ is a classifier, and $\circ$ denotes the composition operation. The feature extractor $f_S$ converts a sample $x$ into a $q$-dimensional feature vector $f_S(x)$, while the classifier $g_S$ assigns class scores to $f_S(x)$, with $(g_S(f_S(x)))_m$ denoting the score for class $m$. The final linear layer of a neural network, typically viewed as the classifier $g_S$, has the weight matrix $W=[w_1,\dots,w_M]\in\mathbb{R}^{q\times M}$, where $w_m$ is the classifier vector for class $m$.

The aim of FFREEDA is to evolve from the source model $h_S$ to \blue{learn multiple personalized classification models $h_k = g_k \circ f_k$, each tailored to the respective client distributions $\mathcal{P}_k(\mathcal{X})$, while encouraging collaboration among clients.} 
In each communication round $r=0,\dots,R-1$, clients train their models \blue{$h_k$} locally for $E$ epochs and send them to the server for aggregation to obtain global models. \blue{See \cref{table:notation} in App.~\ref{app:notation} for a full list of notation.}

\section{Method}
\label{sec:method}

FedWCA comprises three main components: (1) private and parameter-free client clustering, (2) weighted cluster aggregation, and (3) local adaptation via pseudo-labeling.
In line with prevalent SFDA practices \cite{liang_icml2020_shot,sfda-de,bmd}, we fix the classifier during FL, anticipating that the source model's classifier has developed a domain-independent capability for the task. See App.~\ref{app:pseudo_code} for the pseudo-code of FedWCA.

\subsection{Private and 
Parameter-free Client Clustering}
\label{sec:client_clustering}

To manage domain shifts among clients, our method clusters clients privately and without parameters as a first step, based on locally learned feature extractor parameters $f_k$ by each client. This approach removes the need for server access to raw client data and requires only a one-time initial clustering operation.

Unlike traditional clustered FL methods that require hyperparameters such as the number of clusters, our strategy employs a parameter-free clustering: \textit{First Integer Neighbor indices producing a Clustering Hierarchy} (FINCH) \cite{finch}. We use cosine similarity between clients' model parameters to identify the \blue{nearest neighbor $\kappa_k$ for each client $k$}. The adjacency matrix is defined as $A(k,l) = 1$ if $l = \kappa_k$, $k= \kappa_l$, $\kappa_k = \kappa_l$, and $A(k,l) = 0$ otherwise.
Clients linked by this matrix are grouped into the same cluster.

Given the typically large number of parameters in a feature extractor, using all parameters for distance evaluation can lead to a single cluster due to the curse of dimensionality (see \cref{sec:ablation_clustering}). To mitigate this, we focus on the shallow layers where domain knowledge is stored \cite{gromov2024unreasonableineffectivenessdeeperlayers}. 
Specifically, we only use the first-layer parameters as this layer, the furthest from the common classifier among clients, best represents domain-specific information (\cref{sec:ablation_clustering}).
Consequently, the server can derive \blue{$C$ clusters (indexed by $c$) and generate cluster models $f_c$ by averaging the local models in each cluster. In the next section, we construct each client's initial model for the next round using the cluster models.}

\subsection{Weighted Cluster Aggregation}
\label{aggregation_method}

\begin{figure*}[t]
    \centering
    \includegraphics[width=0.95\linewidth]{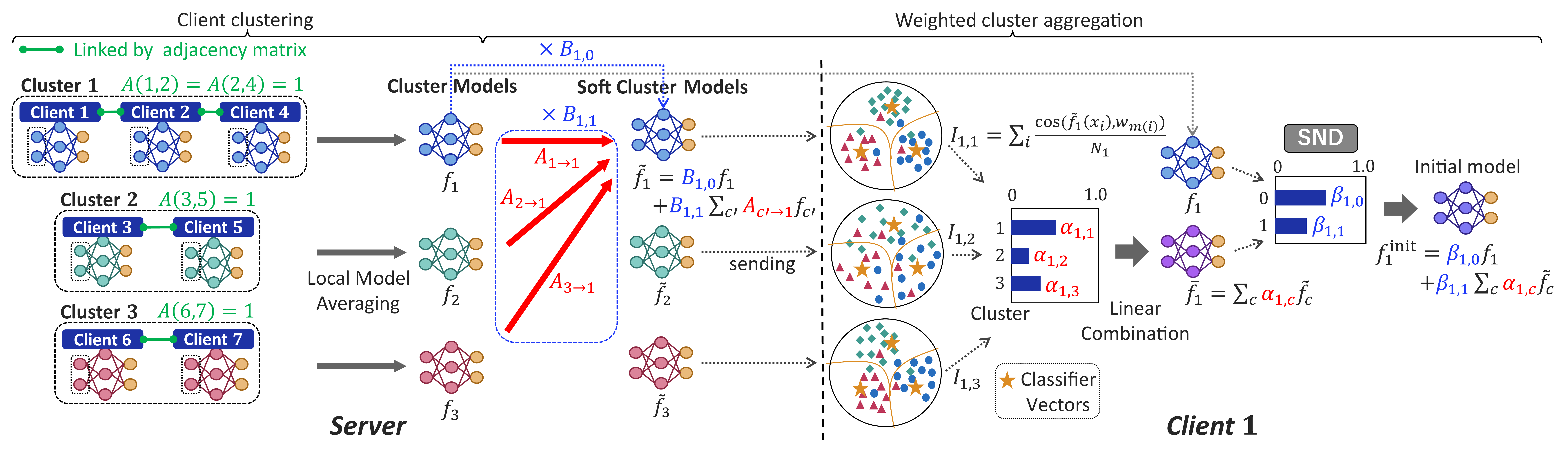}
    \caption{\textbf{Weighted cluster aggregation (WCA).} Our WCA consists of server-side (left) and client-side (right) operations. 
    The client-side operations are represented by client $1$ in cluster $1$. Firstly, the server forms cluster models by averaging the local models within each cluster and develops soft cluster models by integrating models from other clusters. Next, the client merges these soft cluster models utilizing cluster weights determined by how each feature extractor aligns each sample with the respective classifier vector. Additionally, the client combines the merged model with the corresponding cluster model using Soft Neighborhood Density (SND) weights \cite{snd}.
    }
    \label{fig:wca}
\end{figure*}


The core of our FedWCA method is the weighted cluster aggregation, depicted in \cref{fig:wca}. 
\blue{In conventional clustered FL methods, the server distributes the cluster model $f_{c_k}$ to client $k$, where $c_k$ is the cluster index assigned to client $k$, for use as the initial model for the next round.}
This approach limits clients to knowledge solely from their assigned cluster, challenging clients in clusters with few data points or whose domains are significantly different from the source domain. 
To overcome this, our strategy enables clients to blend all cluster models \blue{$f_c$} using specific cluster weights \blue{$\bm{v}_k=(v_{k,1},\dots,v_{k,C})$ (summing to $1$), to form an initial model $f_k^{\rm init} = \sum_{c}v_{k,c}f_c$ in each round} $r$\footnote{In the initial round, \blue{$f_k^{\rm init}$} aligns with the source feature extractor $f_S$, and \blue{$g_k^{\rm init}$} remains fixed to \blue{the source classifier} $g_S$ across all rounds.}.
This blended model not only leverages cross-cluster knowledge but also corrects possible cluster assignment errors. At the end of the training, each client uses its cluster's model for the performance evaluation.
The specific cluster weights calculation method is described below.

\textbf{Local calculation of cluster weights.} 
Firstly, we explore a method for clients to compute cluster weights locally, vital for model personalization. The underlying principle is to assign higher weights to those cluster models that are most beneficial for a client's  domain.

Considering that an effective feature extractor converges samples towards a \blue{corresponding} class-specific classifier vector defined in \cref{sec:problem_formulation}, which effectively acts as a prototype in the feature space, we introduce new metrics \blue{$\bm{\alpha}_k=(\alpha_{k,1},\dots,\alpha_{k,C})$} to represent the cluster weights \blue{$\bm{v}_k$} for each client $k$.
First, client $k$ evaluates the cosine similarity \blue{$\cos(f_c(x_i), w_{m(i)})$} between the feature extractor output \blue{$f_c(x_i)$} and the corresponding class' classifier vector \blue{$w_{m(i)}$, where $m(i) = \argmax_m \cos(f_c(x_i), w_m)$ represents the class whose classifier vector is closest to $f_c(x_i)$}. The similarity scores across all samples are averaged to compute \blue{$I_{k,c} = \sum_{i=1}^{N_k} \cos(f_c(x_i), w_{m(i)})/N_k$} for each cluster $c$ and \blue{$\bm{\alpha}_k$} are derived \blue{using a temperature parameter $T_a$ as follows}:
\begin{gather}
\label{cluster_weights}
    \blue{(\alpha_{k,1}, \dots, \alpha_{k,C}) = \mathrm{softmax}({I_{k,1}/T_a}, \dots, {I_{k,C}/T_a}),}
\end{gather}
\blue{where `${\rm softmax}$' is the softmax function.}

However, this metric focuses solely on the benefits to the individual client, without considering the overall advantages across the clients. Such a client-centric approach may not necessarily contribute positively to the other clients.

\textbf{Global and local calculation of cluster weights.} We calculate the cluster weights, focusing on both individual client benefits and collective advantages. This involves using both cluster models \blue{$f_c$} and `soft cluster models' \blue{$\tilde{f}_c$}, which integrate benefits from other clusters.

For creating soft cluster models, we define a benefit metric $A_{c'\rightarrow c}$, with $\sum_{c'}A_{c'\rightarrow c}=1$, indicating the relative advantage of cluster $c$ for clients in cluster $c'$. \blue{This metric formulates new cluster models by weighting $f_{c'}$ as $\sum_{c'}A_{c'\rightarrow c}f_{c'}$}. To prevent loss of knowledge from the original cluster models, we introduce coefficients \blue{$B_{c,0}$} and \blue{$B_{c,1}$}, summing to 1, for each cluster. These coefficients balance the emphasis between the original cluster model $f_c$ and the newly combined model \blue{$\sum_{c'}A_{c'\rightarrow c}f_{c'}$}. The soft cluster models are thus given by 
\begin{align}
    \blue{\tilde{f}_c = B_{c,0}f_c + B_{c,1}\sum_{c'}A_{c'\rightarrow c}f_{c'}.}
\end{align}
$A_{c'\rightarrow c}$ and $B_{i,c}$ are estimated later in this subsection.

Upon receiving the original cluster model \blue{$f_{c_k}$} and all soft cluster models \blue{$\tilde{f}_c$} from the server, each client $k$ \blue{locally} calculates cluster weights \blue{$\bm{\alpha}_k$} for the soft cluster models, leading to a composite model \blue{$\bar{f}_k=\sum_{c}\alpha_{k,c}\tilde{f}_c$}. To adequately incorporate the original cluster model \blue{$f_{c_k}$}, client $k$ employs pseudo-performance weights \blue{$\bm{\beta}_k = (\beta_{k,0}, \beta_{k,1})$} (summing to 1) to balance between \blue{$f_{c_k}$} and \blue{$\bar{f}_k$}. These weights are derived using Soft Neighborhood Density (SND) \cite{snd}, a method that assesses the overall model efficacy on unlabeled data by analyzing the entropy of similarity distributions between outputs:
\begin{align}
    \label{snd_weight}
    \blue{(\beta_{k,0}, \beta_{k,1})
    = \mathrm{softmax}(S(h_{c_k})/T_b, \ S(\bar{h}_k)/T_b),}
\end{align}
where $T_b$ is the temperature parameter, $S$ is the function to measure SND (detailed in App.~\ref{app:SND}), \blue{$h_{c_k} = g_S \circ f_{c_k}$ , and $\bar{h}_k = g_S \circ \bar{f}_k$}. The initial model for client $k$ is therefore computed as \blue{$f_k^{\rm init} = \beta_{k,0} f_{c_k} + \beta_{k,1}\sum_{c}\alpha_{k,c}\tilde{f}_c$.} \blue{Expanding this equation with respect to the cluster model $f_c$ yields the final cluster weights $\bm{v}_k$, shown in App.~\ref{app:specific_cluster_weights}.}

On the server side, the global coefficients $A_{c \rightarrow c'}$ and \blue{$B_{c,i}$} can be estimated by averaging the local weights \blue{$\alpha_{k,c'}$ and $\beta_{k,i}$} from the previous round within each cluster $c$ because each cluster comprises clients from similar domains. This ensures the cluster weights in \blue{$f_k^{\rm init}$} reflect both local preferences and global trends, aligning individual client needs.

Further discussion on privacy concerns and additional costs in distributing all soft cluster models to each client is found in App.~\ref{app:limitation}. 
It also provides a proposal for reducing communication, computation, and storage costs of our approach by shifting client-side computations to the server.

\subsection{Local Adaptation with Pseudo-Labeling}
\label{sec:local_adaptation_module}

\begin{figure}[t]
    \centering
    \includegraphics[width=0.9\linewidth]{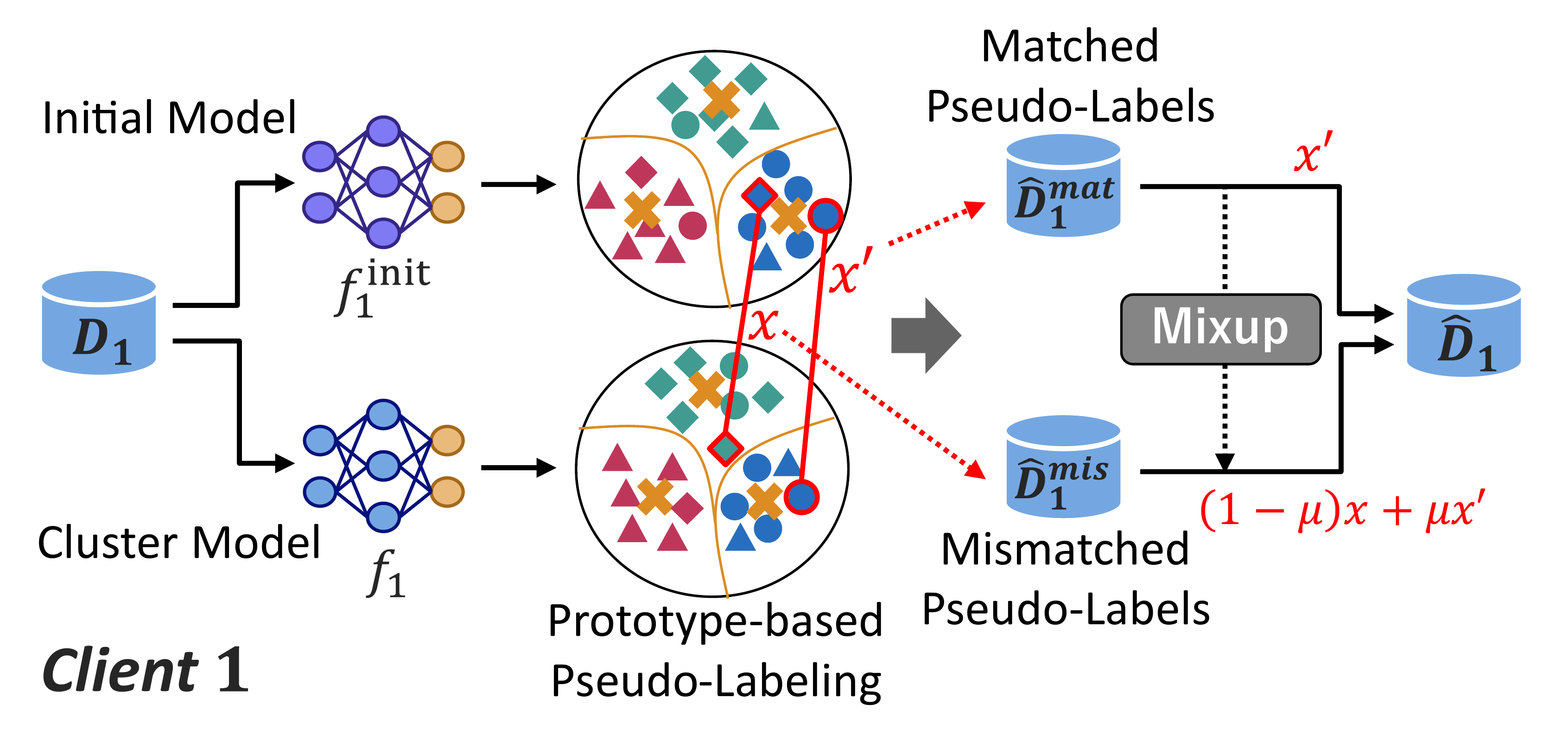}
    \caption{
    \textbf{Pseudo-labeling strategy.} 
    The operation is signified by client $1$ in cluster $1$. Our strategy involves two phases: prototype-based pseudo-labeling with initial and cluster models, and mixup to generate samples with more reliable pseudo-labels. The first phase selects pseudo-labels closer to prototypes, while the second mixes samples with mismatched and matched pseudo-labels.
    }
    \label{fig:pseudo-labeling}
\end{figure}

To adapt \blue{the initial model $f_k^{\rm init}$ constructed in the previous section} to each client \blue{$k$} from only unlabeled data, our local adaptation module adopts the pseudo-labeling method for its simplicity and effectiveness (\cref{fig:pseudo-labeling}), while any existing SFDA method could be applicable for local adaptation.
This choice is particularly pertinent given the limited computational resources and memory capacities of individual clients, compared with other SFDA methods \cite{NEURIPS2021_f5deaeea,yang2022attracting,vdm,cpga,Zhang_2023_CVPR,NEURIPS2022_215aeb07}. 
We further alter an existing feature prototype-based pseudo-labeling strategy \cite{liang_icml2020_shot,dipe,sfda-de,bmd} to lessen overfitting to sparse client data and forgetting global knowledge, and acquire more reliable pseudo-labels. Overfitting is mitigated by using round-fixed pseudo-labels while reliable pseudo-labels are obtained by using both initial and cluster models for pseudo-labeling, followed by mixup. 

\textbf{Pseudo-labeling strategy.} First, we describe the existing prototype-based pseudo-labeling via a model $h=g \circ f$. The procedure involves calculating class-wise feature prototypes $p_m$ for each class $m$ and subsequently determining pseudo-labels $\hat{y}$ for data points $x$ using the cosine similarity:
\begin{align}
\label{prototype}
    p_m = \frac{\sum_{x\in \blue{D_k}}(h(x))_m f(x)}{\sum_{x\in \blue{D_k}}(h(x))_m}, \quad
    \hat{y}=\argmax_{m}d_m,
\end{align}
where $d_m \coloneq \cos(f(x), p_m)$ and $\cos(\cdot,\cdot)$ represents the cosine similarity.
Clients then recompute the prototypes using these one-hot pseudo-labels instead of $h(x)$ in \cref{prototype} and obtains the final pseudo-labels using these prototypes. 

In our setting, each client obtains \blue{$f_k^{\rm init}$} and \blue{$f_{c_k}$} in each round, allowing every data point $x$ to have two pseudo-labels generated by these models. We select the pseudo-label with the larger similarity $d_{\hat{y}}=\cos(f(x), p_{\hat{y}})$. However, as this similarity is calculated on the feature space constructed by each feature extractor, a similarity normalized by the mean similarity between prototypes 
$\sum_{m \ne m'}\cos(p_m,p_{m'})/M(M-1)$
is used to select the correct pseudo-label. This results in a pseudo-labelled dataset \blue{$\hat{D}_k=\{x_i,\hat{y}_i\}_{i=1}^{N_k}$} for each client. Contrary to common practices in SFDA, where pseudo-labels are updated in each epoch, our method keep these pseudo-labels constant throughout each round, ensuring the preservation of the global knowledge in $f_k^{\rm init}$ and $f_{c_k}$.

If \blue{$f_k^{\rm init}$} and \blue{$f_{c_k}$} produce identical pseudo-labels, these are deemed more reliable; distinct pseudo-labels indicate reduced reliability. Therefore, we divide \blue{$\hat{D}_k$} into \blue{$\hat{D}_k^{mat}$} and \blue{$\hat{D}_k^{mis}$}, having matched and mismatched pseudo-labels between \blue{$f_k^{\rm init}$} and \blue{$f_{c_k}$}, respectively. For each instance $(x, \hat{y}) \in \blue{\hat{D}_k^{mis}}$, we mix $x$ with the randomly sampled data point $(x', \hat{y}) \in \blue{\hat{D}_k^{mat}}$, resulting in $x_{mix} = (1-\mu)x + \mu x'$ with the more reliable pseudo-label $\hat{y}$. Here, $\mu$ is a balancing parameter fixed to $0.55$ in our experiments. We thus generate a pseudo-labeled dataset \blue{$\hat{D}_k = \hat{D}_k^{mat} \cup \hat{D}_k^{mix}$}, where \blue{$\hat{D}_k^{mix}$} is a dataset derived from the mixup operation.

\begin{table*}[t]
    \caption{\textbf{Results (Digit-Five, PACS, and Office-Home).} Each row represents the selected source domain denoted by the initial two letters for each dataset, and the numbers are the mean values $\pm$ standard deviations of averaged accuracy (\%) across all clients and all target domains. Thees figures are computed from 15 trials. The best numbers in each row are denoted in bold and the second best are 
    underlined. 
    } 
    \label{tab:main_results}
    \centering
    \small
    \begin{tabular}{cccccccc}\hline
        \multirow{2}{*}{\textbf{Datasets}} &  & \multicolumn{6}{c} {\textbf{Methods}} \\
        \cline{3-8}
        & & Source Only & Local & FedAvg & LADD & FedPCL+PL & FedWCA (ours) \\
        \hline
        \multirow{6}{*}{\textbf{Digit-Five}} & MN & 41.81 $\pm$ 0.46 & 50.59 $\pm$ 1.25 & \underline{55.88 $\pm$ 2.59} & 49.00 $\pm$ 0.82 & 47.84 $\pm$ 0.79 & \textbf{72.74} $\pm$ \textbf{3.57} \\
        & SV & 57.40 $\pm$ 1.39 & 83.00 $\pm$ 0.99 & \underline{83.76 $\pm$ 1.97} & 78.40 $\pm$ 0.64 & 77.73 $\pm$ 0.56 & \textbf{90.13} $\pm$ \textbf{2.08} \\
        & MN-M & 59.87 $\pm$ 1.71 & 69.44 $\pm$ 0.97 & \underline{80.14 $\pm$ 0.46} & 72.93 $\pm$ 1.25 & 70.06 $\pm$ 1.94 & \textbf{82.93} $\pm$ \textbf{0.79}  \\
        & US & 46.93 $\pm$ 0.54 & 56.76 $\pm$ 1.13 & 53.23 $\pm$ 0.73 & \underline{56.93 $\pm$ 1.59} & 56.41 $\pm$ 0.62 & \textbf{58.56} $\pm$ \textbf{4.26} \\
        & SY & 68.69 $\pm$ 0.27 & 81.48 $\pm$ 0.32 & 80.81 $\pm$ 0.98 & \underline{83.79 $\pm$ 0.66} & 81.51 $\pm$ 0.33 & \textbf{84.94} $\pm$ \textbf{1.15} \\
        \cline{2-8}
        & \textbf{Avg.} & 54.94 $\pm$ 9.92 & 68.25 $\pm$ 13.07 & \underline{70.76 $\pm$ 13.49} & 67.67 $\pm$ 13.25 & 66.71 $\pm$ 12.88 & \textbf{77.86} $\pm$ \textbf{11.56} \\
        \hline 
        \multirow{5}{*}{\textbf{PACS}} & Ar & 65.87 $\pm$ 1.19 & 73.01 $\pm$ 1.40 & \underline{76.57 $\pm$ 1.38} & 74.20 $\pm$ 1.48 & 73.83 $\pm$ 2.84 & \textbf{80.63} $\pm$ \textbf{2.44} \\
        & Ca & 68.66 $\pm$ 1.02 & 77.27 $\pm$ 1.23 & 76.15 $\pm$ 0.72 & 77.37 $\pm$ 1.12 & \underline{79.18 $\pm$ 1.51} & \textbf{83.18} $\pm$ \textbf{0.86} \\
        & Ph & 54.20 $\pm$ 1.77 & 63.85 $\pm$ 1.38 & 60.94 $\pm$ 0.96 & 64.77 $\pm$ 1.80 & \underline{65.27 $\pm$ 1.07} & \textbf{65.50} $\pm$ \textbf{1.24} \\
        & Sk & 50.07 $\pm$ 3.21 & 75.56 $\pm$ 6.05 & \underline{80.42 $\pm$ 6.63} & 61.82 $\pm$ 4.88 & 59.41 $\pm$ 6.22 & \textbf{84.22 $\pm$ 6.60} \\
        \cline{2-8}
        & \textbf{Avg.} & 59.70 $\pm$ 8.18 & 72.42 $\pm$ 6.10 & \underline{73.52 $\pm$ 8.23} & 69.54 $\pm$ 7.02 & 69.42 $\pm$ 8.42 & \textbf{78.38 $\pm$ 8.38} \\
        \hline
        \multirow{5}{*}{\textbf{Office-Home}} & Ar & 54.03 $\pm$ 0.30 & 62.50 $\pm$ 0.78 & \underline{63.81 $\pm$ 0.85} & 57.33 $\pm$ 0.72 & 58.20 $\pm$ 0.73 & \textbf{66.06 $\pm$ 0.50} \\
        & Cl & 58.36 $\pm$ 1.29 & 64.57 $\pm$ 0.70 & \textbf{68.73 $\pm$ 0.75} & 61.72 $\pm$ 0.61 & 62.39 $\pm$ 0.98 & \underline{68.32 $\pm$ 0.42} \\
        & Pr & 55.36 $\pm$ 0.72 & 59.17 $\pm$ 0.40 & \underline{60.15 $\pm$ 0.48} & 57.22 $\pm$ 0.76 & 58.85 $\pm$ 0.46 & \textbf{61.46 $\pm$ 0.88} \\
        & Re & 62.28 $\pm$ 0.35 & 65.38 $\pm$ 1.15 & \underline{65.93 $\pm$ 0.34} & 64.23 $\pm$ 0.44 & 65.75 $\pm$ 0.80 & \textbf{68.06 $\pm$ 0.46} \\
        \cline{2-8}
        & \textbf{Avg.} & 57.51 $\pm$ 3.33 & 62.90 $\pm$ 2.54 & \underline{64.65 $\pm$ 3.22} & 60.25 $\pm$ 2.96 & 61.30 $\pm$ 3.14 & \textbf{65.97 $\pm$ 2.83} \\
        \hline
    \end{tabular}
    \label{table:full_pacs}
\end{table*}

\textbf{Loss function.} 
We adopt the same loss function as SHOT \cite{liang_icml2020_shot} because of its simplicity.
The total loss function $\blue{\mathcal{L}_k}=\mathcal{L}_{\rm IM} + \lambda\mathcal{L}_{\rm CE}$ for client $k$ with a balancing parameter $\lambda$ comprises the cross-entropy loss for the pseudo-labeled dataset \blue{$\hat{D}_k$}: $\mathcal{L}_{\rm CE}$ and an Information Maximization (IM) loss \cite{im_loss,liang_icml2020_shot} for the unlabeled dataset \blue{$D_k$}: $\mathcal{L}_{\rm IM}$ (detailed in App.~\ref{app:loss}). 
The IM loss promotes confident model outputs and discourages trivial score distributions.

\section{Experiment}
\label{sec:experiment}

\textbf{Datasets.} Our method was evaluated on three classification datasets, each with diverse domains. The \textbf{Digit-Five} dataset encompasses 5 benchmarks for handwritten-digit classification across 10 categories. It includes MNIST\cite{lecun_ieee1998_lenet}, MNIST-M\cite{Ganin_2017_mnistm}, SVHN\cite{netzer_2011_svhn}, SYNTH\cite{Ganin_2017_mnistm}, and USPS. The \textbf{PACS} \cite{li_ieee2017_pacs} dataset consists of 4 domains with 7 categories each, including Art Painting, Cartoon, Photo, and Sketch. The \textbf{Office-Home} \cite{office-home} dataset is designed for object recognition, comprising 4 domains with 65 categories each and includes Art, Clipart, Product, and Real-World.
For each dataset, one domain was selected as the source, and the remaining domains served as target domains. Data samples from each target domain were distributed to 8 clients for Digit-Five and 3 clients for PACS and Office-Home per domain, all in an i.i.d. manner. This resulted in a total of 32 clients for Digit-Five and 9 for PACS and Office-Home.

\textbf{Models.} 
Our model architecture reflects common structures used in SFDA research like SHOT \cite{liang_icml2020_shot}, incorporating a feature extractor, a dimension-reducing bottleneck layer, and a classifier. The feature extractor differs across datasets. Following prior domain adaptation and FL studies, we use a randomly initialized LeNet \cite{lecun_ieee1998_lenet} with two convolution layers for Digit-Five. For PACS and Office-Home, we use ResNet-18 and ResNet-50 \cite{he_ieee2016_resnet} respectively, both pre-trained on ImageNet. The classifier is the last linear layer and the bottleneck layer consists of a fully connected layer equipped with batch normalization \cite{batch_norm} and dropout \cite{dropout} for all datasets.

\textbf{Baselines.} For comparative analysis, we benchmark our method \textbf{FedWCA} against several baseline models. \textbf{Source Only} uses only pre-trained source models. In terms of SFDA approaches, \textbf{Local} and \textbf{FedAvg} apply SFDA methods to client-side learning. While `Local' represents a non-federated approach, `FedAvg' incorporates model aggregation by FedAvg\cite{mcmahan_pmlr2017_fedavg} at the server. 
As a foundational SFDA method, SHOT \cite{liang_icml2020_shot} is used in these approaches, with round-fixed pseudo-labels for FedAvg,
akin to our method. We also assess against \textbf{LADD} \cite{shenaj_wacv2023_ladd}, an existing FFREEDA method.
\blue{To examine if existing FL approaches for non-i.d.d. data are applicable to FFREEDA by adding a pseudo-labeling step, we compare our method with \textbf{FedPCL+PL}. 
This method integrates FedPCL \cite{fedpcl} with prototype-based pseudo-labeling. FedPCL aligns with our setting as it adapts pre-trained models to clients.} 
\blue{See App.~\ref{app:additional_comparison} for comparisons with other FL methods extended to FFREEDA.}



\begin{figure*}[t]
  \centering
  \begin{subfigure}{1.0\linewidth}
    \centering
    \includegraphics[width=0.75\linewidth]{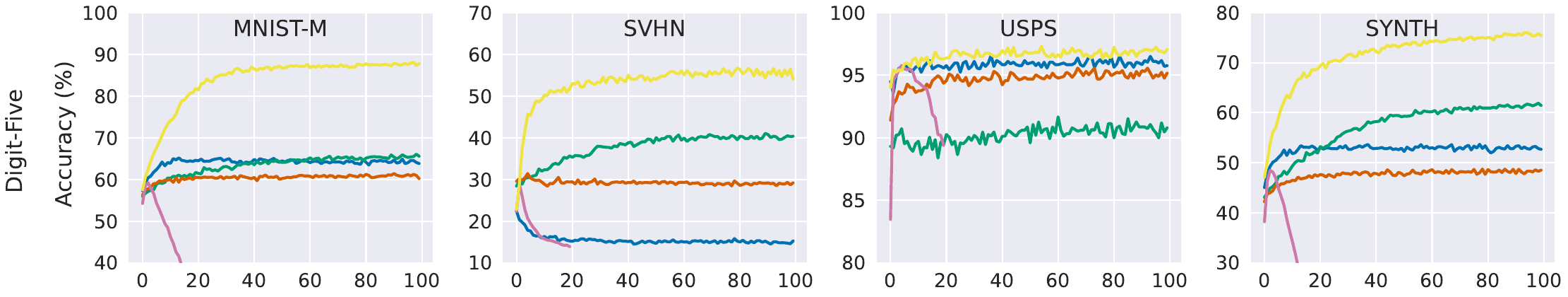}
  \end{subfigure}
  \hfill
  \begin{subfigure}{1.0\linewidth}
    \centering
    \includegraphics[width=0.68\linewidth]{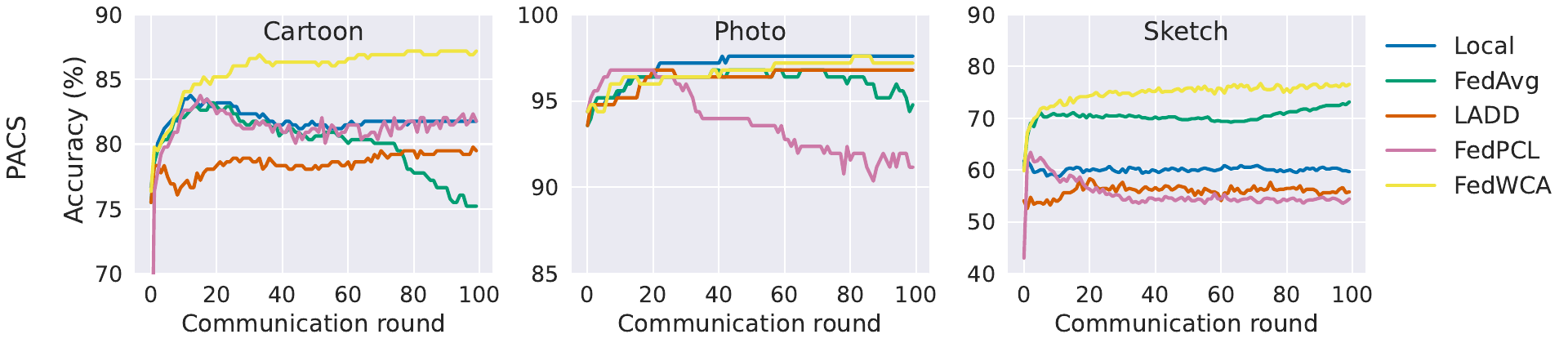}
  \end{subfigure}
  \caption{\textbf{Comparison of average accuracy over clients within each target domain.}
  The source domain is MNIST for Digit-Five (top) and Art Painting for PACS (bottom). The target domains are the others.
  FedWCA considerably enhances the accuracy of clients in specific target domains such as SVHN for Digit-Five and Sketch for PACS, while maintaining the performance of other domains.
  }
  \label{fig:accuracy_plot}
\end{figure*}

\textbf{Evaluation.} 
We trained five source models using standard cross-entropy loss without incorporating any target data information to ensure a fair comparison. We then conducted three trials for each method per source model, totaling fifteen trials. The evaluation metric is the average accuracy across all clients, assessed by the final personalized models using client-specific test data. The results provide the mean values of this average accuracy over all trials.

See App.~\ref{app:experimental_details} for more detailed experimental settings.

\subsection{Comparison with Baselines}
\label{sec:results}

The comparison across the three datasets (\cref{tab:main_results}) reveals varied performances. The `Source Only' approach demonstrates limited effectiveness, while the `Local' approach, applying SFDA to local learning, significantly enhances performance. `FedAvg' further improves upon `Local', showcasing the benefits of FL. However, `FedAvg' experiences performance drops in some cases due to domain shift, such as a decrease from 56.76\% to 53.23\% on USPS for Digit-Five.
`LADD' and `FedPCL+PL' mitigate domain shift issues somewhat and improve accuracy in some source domains, but are inferior to `FedAvg' in many domains. This is because LADD's primary focus is not classification tasks and therefore it uses a simple pseudo-labelling strategy, and furthermore it does not fully utilize knowledge from all domains. On the other hand, FedPCL originally employs true labels at every step and substituting them all with pseudo-labels could compromise performance.

Our FedWCA outperforms the best compared methods by notable margins on almost all source domains (7.10\% for Digit-Five, 4.86\% for PACS, 1.32\% for Office-Home on average). The results also confirm that FedWCA statistically outperforms the others in a majority of source domains.
The improvement on Office-Home is more modest, likely due to the smaller domain gap in this dataset, which can be judged from the fact that the improvement for `Source Only' by `Local' is considerably smaller (5.39\%) than for the other datasets (13.31\% for Digit-Five and 12.72\% for PACS). 

In \cref{fig:accuracy_plot}, we analyze the changes in average accuracy of clients in each target domain, using specific source domains, such as MNIST for Digit-Five (top) and Art Painting for PACS (bottom). Unlike `FedAvg', FedWCA significantly improves accuracy in almost all target domains without compromising accuracy in specific target domains such as USPS for Digit-Five and Photo for PACS.

\begin{table}[t]
    \centering
    \caption{\textbf{Ablation on cluster weights calculation.} The FedWCA's cluster weights calculation is replaced by various methods such as one-hot weights (One-hot), equal weights (Equal), a hybrid of them (One-Equal), an adaptive version of One-Equal (One-Equal-A), and a local calculation-only FedWCA variant described in \cref{aggregation_method} (FedWCA-L). The original FedWCA surpasses all other methods across all PACS source domains.}%
    \small
    \begin{tabular}{rccccc}\hline
        \multirow{2}{*}{\textbf{Methods}} &  \multicolumn{5}{c} {\textbf{PACS}} \\
        \cline{2-6}
        & Ar & Ca & Ph & Sk & \textbf{Avg.} \\ \hline
        One-hot & 73.22 & 77.34 & 64.45 & 80.40 & 73.85 \\
        Equal & 76.91 & 76.03 & 61.16 & 79.55 & 73.41  \\
        One-Equal & 77.89 & 79.69 & 64.76 & 83.13 & 76.37 \\
        One-Equal-A & 77.80 & 79.17 & 63.64 & 83.29 & 75.98  \\
        FedWCA-L & 77.12 & 81.94 & 62.25  & 82.31 & 75.91 \\
        FedWCA & \textbf{80.63} & \textbf{83.18} & \textbf{65.50}  & \textbf{84.22} & \textbf{78.38} \\
        \hline
    \end{tabular}
    \label{table:pacs_ablation_wca}
\end{table}

\subsection{Ablation Studies}
\label{sec:ablation_study}

\subsubsection{Cluster Weight Calculation}

To validate the effectiveness of our cluster weight 
calculation method, we conducted an ablation study comparing our method with simpler cluster weight computation methods including: (1) \textbf{One-hot}: Assigns a weight of 1 to each client's own cluster and 0 to others.
(2) \textbf{Equal}: Allocates an equal weight to all clusters, resembling FedAvg.
(3) \textbf{One-Equal}: A hybrid approach that assigns a weight of $p$ (set to 0.8) to each client's cluster and distributes the remaining weight equally among
other clusters.
(4) \textbf{One-Equal-A}: An adaptive method that combines models created by 
One-hot and Equal using SND, as in FedWCA.
(5) \textbf{FedWCA-L}: A variant of FedWCA wherein cluster weights are computed solely locally, described in \cref{aggregation_method}.

The performance comparison of these methods on PACS is shown in \cref{table:pacs_ablation_wca}. 
The hybrid approaches, One-Equal and its adaptive variant, and FedWCA-L, increase accuracy compared to simpler methods like One-hot and Equal. FedWCA, however, outperforms them across all source domains. The superior performance illustrates the benefits of our method that integrates global and local insights in cluster weight calculation.
See App.~\ref{app:cluster_weights} for examples of specific cluster weights in our methods.

\subsubsection{Client Clustering Method}
\label{sec:ablation_clustering}

\begin{table}[t]
    \centering
    \caption{
    \textbf{Ablation on layers used for clustering.} 
    Art Painting of PACS is used as the source domain, and ResNet-18 is employed. Clients $C, C', C''$ belong to Cartoon, $P, P', P''$ to Photo, and $S, S', S''$ to Sketch. The table below reports the cluster IDs (denoted by $\bigcirc$, $\triangle$, and $\square$) assigned to each client. 
    Ideally, each client group $(C, C', C'')$, $(P, P', P'')$, and $(S, S', S'')$ should be grouped together, as is achieved by the first and second layers.
    }

    \label{table:clustering_result}
    \footnotesize
    \begin{tabular}{rccccccccc}\hline
        \textbf{ } &  \multicolumn{9}{c} {\textbf{Clients}} \\
        \cline{2-10}
        \textbf{Layer} & $C$ & $C'$ & $C''$ & $P$ & $P'$ & $P''$ & $S$ & $S'$ & $S''$ \\ \hline
        1st & $\bigcirc$ & $\bigcirc$ & $\bigcirc$ & $\triangle$ & $\triangle$ & $\triangle$ & $\square$ & $\square$ & $\square$ \\
        2nd & $\bigcirc$ & $\bigcirc$ & $\bigcirc$ & $\triangle$ & $\triangle$ & $\triangle$ & $\square$ & $\square$ & $\square$ \\
        3rd & $\bigcirc$ & $\bigcirc$ & $\bigcirc$ & $\bigcirc$ & $\bigcirc$ & $\bigcirc$ & $\bigcirc$ & $\bigcirc$ & $\bigcirc$ \\
        4th & $\bigcirc$ & $\bigcirc$ & $\bigcirc$ & $\bigcirc$ & $\bigcirc$ & $\bigcirc$ & $\triangle$ & $\triangle$ & $\triangle$ \\
        17th & $\bigcirc$ & $\bigcirc$ & $\bigcirc$ & $\bigcirc$ & $\bigcirc$ & $\bigcirc$ & $\bigcirc$ & $\bigcirc$ & $\bigcirc$ \\
        All & $\bigcirc$ & $\bigcirc$ & $\bigcirc$ & $\bigcirc$ & $\bigcirc$ & $\bigcirc$ & $\bigcirc$ & $\bigcirc$ & $\bigcirc$ \\
        \hline
    \end{tabular}
\end{table}

\cref{table:clustering_result} evaluates the clustering performance using different layers' parameters in ResNet-18, showing that using only the first or second layer is the best practice (see App.~\ref{app:clustering} for full results).
The results indicate that clustering performance diminishes as a deeper layer rather than the first layer (\eg, 17th) is used. Using all layers leads to a single cluster (indexed by $\bigcirc$ in \cref{table:clustering_result}) due to the curse of dimensionality.

Despite successful clustering in PACS and Office-Home, some errors were observed in Digit-Five (see App.~\ref{app:clustering}). 
Nonetheless, our method improves performance even with incorrect cluster assignments by effectively combining cluster models with individualized weights (App.~\ref{app:clustering}).



\subsubsection{Pseudo-labeling Strategy}


We examine the contribution of each key component of our pseudo-labeling strategy. The pseudo-labeling strategy of FedWCA consists of four key components (see \cref{sec:local_adaptation_module}): prototype-based pseudo-labeling (PP), fixed pseudo-labels (FP) in each round, two-models-based pseudo-labeling (TM), which utilizes both the initial and original cluster models, and mixup (MU) between the samples with the matched and mismatched pseudo-labels.

To assess the effectiveness of each component, we incrementally activated them in FedWCA and observed their impact on performance in the PACS dataset. The results, presented in \cref{table:pacs_ablation_local}, reveal that the full integration of all components leads to the highest performance across all source domains. This finding underscores the significance of each element, with every component contributing meaningfully to the overall efficacy of the method.

The above results demonstrate that PP and FP independently enhance accuracy and may also be incorporated into LADD. Similarly, the IM loss adopted by our method could contribute to LADD's expansion, tailoring it more towards the classification task, akin to the local adaptation module with `FedAvg' in our experiments. \cref{tab:extended_ladd} compares the performance of the extended LADD with our FedWCA. A comparison between \cref{tab:main_results} and \cref{tab:extended_ladd} reveals a significant performance boost in the extended LADD over the original. However, FedWCA outperforms this improved LADD, further supporting our method's efficiency.

\begin{table}[t]
    \caption{\textbf{Ablation on pseudo-labeling strategy}. Key components of FedWCA are \textbf{PP}: prototype-based pseudo-labeling, \textbf{FP}: fixed pseudo-labels in each round, \textbf{TM}: two-models-based pseudo-labeling using initial and cluster models, and \textbf{MU}: mixup. All components prove effective in most PACS source domains.}
    \centering
    \footnotesize
    \begin{tabular}{ccccccccc}\hline
        \multirow{2}{*}{PP} & \multirow{2}{*}{FP} & \multirow{2}{*}{TM} & \multirow{2}{*}{MU} & \multicolumn{5}{c} {\textbf{PACS}} \\
        \cline{5-9}
        & & & & Ar & Ca & Ph & Sk & \textbf{Avg.} \\ \hline
        & \checkmark & & &  72.98 & 76.44 & 61.78 & 68.73 &	70.29 \\
        \checkmark & & & &  77.63 & 75.66 & 65.48 & 81.31 & 74.39 \\
        \checkmark & \checkmark &  &  & 78.37 & 81.54 & 65.23 & 83.56 & 77.17 \\
        \checkmark & \checkmark & \checkmark & & 78.82 & 81.72 & \textbf{66.27} & 83.92 & 77.68 \\
        \checkmark & \checkmark & \checkmark & \checkmark & \textbf{80.63} & \textbf{83.18} & 65.50 & \textbf{84.22} & \textbf{78.38} \\ \hline
    \end{tabular}
    \label{table:pacs_ablation_local}
\end{table}

\begin{table}[t]
    \caption{\textbf{Comparison with extended LADD.} The prior FFREEDA method LADD is extended by incorporating \textbf{PP}: prototype-based pseudo-labeling, \textbf{FP}: fixed pseudo-labels in each round, and \textbf{IM}: IM loss. Nevertheless, FedWCA outperforms it.} 
    \label{tab:extended_ladd}
    \centering
    \footnotesize
    \begin{tabular}{cccc}\hline
        \multirow{3}{*}{\textbf{Datasets}} &  & \multicolumn{2}{c} {\textbf{Methods}} \\
        \cline{3-4}
        & & LADD+PP+FP+IM & FedWCA \\
        \hline
        \multirow{6}{*}{\textbf{Digit-Five}} & MN & 60.45 $\pm$ 2.14 & \textbf{72.74} $\pm$ \textbf{3.57} \\
        & SV & 90.06 $\pm$ 2.41 & \textbf{90.13} $\pm$ \textbf{2.08} \\
        & MN-M & 72.11 $\pm$ 3.06 & \textbf{82.93} $\pm$ \textbf{0.79}  \\
        & US & 56.96 $\pm$ 1.67 & \textbf{58.56} $\pm$ \textbf{4.26} \\
        & SY & \textbf{87.12 $\pm$ 0.65} & 84.94 $\pm$ 1.15 \\
        \cline{2-4}
        & \textbf{Avg.} & 73.34 $\pm$ 13.71 & \textbf{77.86} $\pm$ \textbf{11.56} \\
        \hline 
        \multirow{5}{*}{\textbf{PACS}} & Ar & 73.19 $\pm$ 1.73 & \textbf{80.63} $\pm$ \textbf{2.44} \\
        & Ca & 75.98 $\pm$ 2.49 & \textbf{83.18} $\pm$ \textbf{0.86} \\
        & Ph & 65.15 $\pm$ 1.27 & \textbf{65.50} $\pm$ \textbf{1.24} \\
        & Sk & 79.24 $\pm$ 6.76 & \textbf{84.22 $\pm$ 6.60} \\
        \cline{2-4}
        & \textbf{Avg.} & 73.39 $\pm$ 6.41 & \textbf{78.38 $\pm$ 8.38} \\
        \hline
        \multirow{5}{*}{\textbf{Office-Home}} & Ar & 65.04 $\pm$ 0.82 & \textbf{66.06 $\pm$ 0.50} \\
        & Cl & 67.88 $\pm$ 0.67 & \textbf{68.32 $\pm$ 0.42} \\
        & Pr & 61.06 $\pm$ 0.53 & \textbf{61.46 $\pm$ 0.88} \\
        & Re & 67.58 $\pm$ 0.80 & \textbf{68.06 $\pm$ 0.46} \\
        \cline{2-4}
        & \textbf{Avg.} & 65.39 $\pm$ 3.14 & \textbf{65.97 $\pm$ 2.83} \\
        \hline
    \end{tabular}
    \label{table:full_pacs}
\end{table}

\section{Conclusion}
This paper introduces FedWCA, a novel method for FFREEDA classification, addressing two problems: lack of labeled data post-pre-training and domain shift among clients. 
Experimental results show FedWCA's superior performance over several baselines, validating its effectiveness in FFREEDA. We discuss the limitations in App.~\ref{app:limitation}.

\section*{Acknowledgments}
This R\&D includes the results of " Research and development of optimized AI 
technology by secure data coordination (JPMI00316)" by the Ministry of Internal Affairs 
and Communications (MIC), Japan.

{\small
\bibliographystyle{ieee_fullname}
\bibliography{egbib}

\begin{thebibliography}{10}\itemsep=-1pt

\bibitem{fl+hc}
Christopher Briggs, Zhong Fan, and Peter Andras.
\newblock Federated learning with hierarchical clustering of local updates to improve training on non-iid data.
\newblock In {\em 2020 International Joint Conference on Neural Networks (IJCNN)}, pages 1--9, 2020.

\bibitem{fedgc}
Debora Caldarola, Massimiliano Mancini, Fabio Galasso, Marco Ciccone, Emanuele Rodolà, and Barbara Caputo.
\newblock Cluster-driven graph federated learning over multiple domains.
\newblock In {\em 2021 IEEE/CVF Conference on Computer Vision and Pattern Recognition Workshops (CVPRW)}, pages 2743--2752, 2021.

\bibitem{uifca}
Jichan Chung, Kangwook Lee, and Kannan Ramchandran.
\newblock Federated unsupervised clustering with generative models.
\newblock In {\em AAAI 2022 International Workshop on Trustable, Verifiable and Auditable Federated Learning}, 2022.

\bibitem{fedrep}
Liam Collins, Hamed Hassani, Aryan Mokhtari, and Sanjay Shakkottai.
\newblock Exploiting shared representations for personalized federated learning.
\newblock In Marina Meila and Tong Zhang, editors, {\em Proceedings of the 38th International Conference on Machine Learning}, volume 139 of {\em Proceedings of Machine Learning Research}, pages 2089--2099. PMLR, 18--24 Jul 2021.

\bibitem{semifl}
Enmao Diao, Jie Ding, and Vahid Tarokh.
\newblock Semifl: Semi-supervised federated learning for unlabeled clients with alternate training.
\newblock In S. Koyejo, S. Mohamed, A. Agarwal, D. Belgrave, K. Cho, and A. Oh, editors, {\em Advances in Neural Information Processing Systems}, volume~35, pages 17871--17884. Curran Associates, Inc., 2022.

\bibitem{sfda-de}
Ning Ding, Yixing Xu, Yehui Tang, Chao Xu, Yunhe Wang, and Dacheng Tao.
\newblock Source-free domain adaptation via distribution estimation.
\newblock In {\em 2022 IEEE/CVF Conference on Computer Vision and Pattern Recognition (CVPR)}, pages 7202--7212, 2022.

\bibitem{perfedavg}
Alireza Fallah, Aryan Mokhtari, and Asuman Ozdaglar.
\newblock Personalized federated learning with theoretical guarantees: a model-agnostic meta-learning approach.
\newblock In {\em Proceedings of the 34th International Conference on Neural Information Processing Systems}, NIPS'20, Red Hook, NY, USA, 2020. Curran Associates Inc.

\bibitem{fedssl}
Chenyou Fan, Junjie Hu, and Jianwei Huang.
\newblock Private semi-supervised federated learning.
\newblock In Lud~De Raedt, editor, {\em Proceedings of the Thirty-First International Joint Conference on Artificial Intelligence, {IJCAI-22}}, pages 2009--2015. International Joint Conferences on Artificial Intelligence Organization, 7 2022.
\newblock Main Track.

\bibitem{kd3a}
Haozhe Feng, Zhaoyang You, Minghao Chen, Tianye Zhang, Minfeng Zhu, Fei Wu, Chao Wu, and Wei Chen.
\newblock Kd3a: Unsupervised multi-source decentralized domain adaptation via knowledge distillation.
\newblock In Marina Meila and Tong Zhang, editors, {\em Proceedings of the 38th International Conference on Machine Learning}, volume 139 of {\em Proceedings of Machine Learning Research}, pages 3274--3283. PMLR, 18--24 Jul 2021.

\bibitem{fruda}
Shaoduo Gan, Akhil Mathur, Anton Isopoussu, Fahim Kawsar, Nadia Berthouze, and Nicholas~D. Lane.
\newblock Fruda: Framework for distributed adversarial domain adaptation.
\newblock {\em IEEE Transactions on Parallel and Distributed Systems}, 33(11):3153--3164, 2022.

\bibitem{Ganin_2017_mnistm}
Yaroslav Ganin, Evgeniya Ustinova, Hana Ajakan, Pascal Germain, Hugo Larochelle, Fran{\c{c}}ois Laviolette, Mario Marchand, and Victor Lempitsky.
\newblock {\em Domain-Adversarial Training of Neural Networks}, pages 189--209.
\newblock Springer International Publishing, Cham, 2017.

\bibitem{ifca}
Avishek Ghosh, Jichan Chung, Dong Yin, and Kannan Ramchandran.
\newblock An efficient framework for clustered federated learning.
\newblock In {\em Proceedings of the 34th International Conference on Neural Information Processing Systems}, NIPS'20, Red Hook, NY, USA, 2020. Curran Associates Inc.

\bibitem{im_loss}
Ryan Gomes, Andreas Krause, and Pietro Perona.
\newblock Discriminative clustering by regularized information maximization.
\newblock In {\em Proceedings of the 23rd International Conference on Neural Information Processing Systems - Volume 1}, NIPS'10, page 775–783, Red Hook, NY, USA, 2010. Curran Associates Inc.

\bibitem{gromov2024unreasonableineffectivenessdeeperlayers}
Andrey Gromov, Kushal Tirumala, Hassan Shapourian, Paolo Glorioso, and Daniel~A. Roberts.
\newblock The unreasonable ineffectiveness of the deeper layers, 2024.

\bibitem{he_ieee2016_resnet}
Kaiming He, Xiangyu Zhang, Shaoqing Ren, and Jian Sun.
\newblock Deep residual learning for image recognition.
\newblock In {\em 2016 IEEE Conference on Computer Vision and Pattern Recognition (CVPR)}, pages 770--778, 2016.

\bibitem{fade}
Junyuan Hong, Zhuangdi Zhu, Shuyang Yu, Zhangyang Wang, Hiroko~H. Dodge, and Jiayu Zhou.
\newblock Federated adversarial debiasing for fair and transferable representations.
\newblock In {\em Proceedings of the 27th ACM SIGKDD Conference on Knowledge Discovery \& Data Mining}, KDD '21, page 617–627, New York, NY, USA, 2021. Association for Computing Machinery.

\bibitem{fedamp}
Yutao Huang, Lingyang Chu, Zirui Zhou, Lanjun Wang, Jiangchuan Liu, Jian Pei, and Yong Zhang.
\newblock Personalized cross-silo federated learning on non-iid data.
\newblock {\em Proceedings of the AAAI Conference on Artificial Intelligence}, 35(9):7865--7873, May 2021.

\bibitem{batch_norm}
Sergey Ioffe and Christian Szegedy.
\newblock Batch normalization: Accelerating deep network training by reducing internal covariate shift.
\newblock In Francis Bach and David Blei, editors, {\em Proceedings of the 32nd International Conference on Machine Learning}, volume~37 of {\em Proceedings of Machine Learning Research}, pages 448--456, Lille, France, 07--09 Jul 2015. PMLR.

\bibitem{fedmatch}
Wonyong Jeong, Jaehong Yoon, Eunho Yang, and Sung~Ju Hwang.
\newblock Federated semi-supervised learning with inter-client consistency \& disjoint learning.
\newblock In {\em International Conference on Learning Representations}, 2021.

\bibitem{efficientfda}
Hua Kang, Zhiyang Li, and Qian Zhang.
\newblock Communicational and computational efficient federated domain adaptation.
\newblock {\em IEEE Transactions on Parallel and Distributed Systems}, 33(12):3678--3689, 2022.

\bibitem{protofl}
Hansol Kim, Youngjun Kwak, Minyoung Jung, Jinho Shin, Youngsung Kim, and Changick Kim.
\newblock Protofl: Unsupervised federated learning via prototypical distillation.
\newblock In {\em 2023 IEEE/CVF International Conference on Computer Vision (ICCV)}, pages 6447--6456, 2023.

\bibitem{fedfmc}
Kavya Kopparapu and Eric Lin.
\newblock Fedfmc: Sequential efficient federated learning on non-iid data, 2020.

\bibitem{lecun_ieee1998_lenet}
Y. Lecun, L. Bottou, Y. Bengio, and P. Haffner.
\newblock Gradient-based learning applied to document recognition.
\newblock {\em Proceedings of the IEEE}, 86(11):2278--2324, 1998.

\bibitem{pmlr-v162-lee22c}
Jonghyun Lee, Dahuin Jung, Junho Yim, and Sungroh Yoon.
\newblock Confidence score for source-free unsupervised domain adaptation.
\newblock In Kamalika Chaudhuri, Stefanie Jegelka, Le Song, Csaba Szepesvari, Gang Niu, and Sivan Sabato, editors, {\em Proceedings of the 39th International Conference on Machine Learning}, volume 162 of {\em Proceedings of Machine Learning Research}, pages 12365--12377. PMLR, 17--23 Jul 2022.

\bibitem{li_ieee2017_pacs}
Da Li, Yongxin Yang, Yi-Zhe Song, and Timothy~M. Hospedales.
\newblock Deeper, broader and artier domain generalization.
\newblock In {\em 2017 IEEE International Conference on Computer Vision (ICCV)}, pages 5543--5551, 2017.

\bibitem{cbafed}
Ming Li, Qingli Li, and Yan Wang.
\newblock Class balanced adaptive pseudo labeling for federated semi-supervised learning.
\newblock In {\em 2023 IEEE/CVF Conference on Computer Vision and Pattern Recognition (CVPR)}, pages 16292--16301, 2023.

\bibitem{fedprox}
Tian Li, Anit~Kumar Sahu, Manzil Zaheer, Maziar Sanjabi, Ameet Talwalkar, and Virginia Smith.
\newblock Federated optimization in heterogeneous networks.
\newblock In I. Dhillon, D. Papailiopoulos, and V. Sze, editors, {\em Proceedings of Machine Learning and Systems}, volume~2, pages 429--450, 2020.

\bibitem{panm}
Zexi Li, Jiaxun Lu, Shuang Luo, Didi Zhu, Yunfeng Shao, Yinchuan Li, Zhimeng Zhang, Yongheng Wang, and Chao Wu.
\newblock Towards effective clustered federated learning: A peer-to-peer framework with adaptive neighbor matching.
\newblock {\em IEEE Transactions on Big Data}, pages 1--16, 2022.

\bibitem{liang_icml2020_shot}
Jian Liang, Dapeng Hu, and Jiashi Feng.
\newblock Do we really need to access the source data? {S}ource hypothesis transfer for unsupervised domain adaptation.
\newblock In Hal~Daumé III and Aarti Singh, editors, {\em Proceedings of the 37th International Conference on Machine Learning}, volume 119 of {\em Proceedings of Machine Learning Research}, pages 6028--6039. PMLR, 13--18 Jul 2020.

\bibitem{rscfed}
Xiaoxiao Liang, Yiqun Lin, Huazhu Fu, Lei Zhu, and Xiaomeng Li.
\newblock Rscfed: Random sampling consensus federated semi-supervised learning.
\newblock In {\em 2022 IEEE/CVF Conference on Computer Vision and Pattern Recognition (CVPR)}, pages 10144--10153, 2022.

\bibitem{Litrico_2023_CVPR}
Mattia Litrico, Alessio Del~Bue, and Pietro Morerio.
\newblock Guiding pseudo-labels with uncertainty estimation for source-free unsupervised domain adaptation.
\newblock In {\em 2023 IEEE/CVF Conference on Computer Vision and Pattern Recognition (CVPR)}, pages 7640--7650, 2023.

\bibitem{orchestra}
Ekdeep Lubana, Chi~Ian Tang, Fahim Kawsar, Robert Dick, and Akhil Mathur.
\newblock Orchestra: Unsupervised federated learning via globally consistent clustering.
\newblock In Kamalika Chaudhuri, Stefanie Jegelka, Le Song, Csaba Szepesvari, Gang Niu, and Sivan Sabato, editors, {\em Proceedings of the 39th International Conference on Machine Learning}, volume 162 of {\em Proceedings of Machine Learning Research}, pages 14461--14484. PMLR, 17--23 Jul 2022.

\bibitem{three}
Yishay Mansour, Mehryar Mohri, Jae Ro, and Ananda~Theertha Suresh.
\newblock Three approaches for personalization with applications to federated learning.
\newblock {\em arXiv preprint arXiv:2002.10619}, 2020.

\bibitem{mcmahan_pmlr2017_fedavg}
Brendan McMahan, Eider Moore, Daniel Ramage, Seth Hampson, and Blaise Aguera~y Arcas.
\newblock {Communication-Efficient Learning of Deep Networks from Decentralized Data}.
\newblock In Aarti Singh and Jerry Zhu, editors, {\em Proceedings of the 20th International Conference on Artificial Intelligence and Statistics}, volume~54 of {\em Proceedings of Machine Learning Research}, pages 1273--1282. PMLR, 20--22 Apr 2017.

\bibitem{pfedmb}
Junki Mori, Tomoyuki Yoshiyama, Ryo Furukawa, and Isamu Teranishi.
\newblock Personalized federated learning with multi-branch architecture.
\newblock In {\em 2023 International Joint Conference on Neural Networks (IJCNN)}, pages 1--8, 2023.

\bibitem{netzer_2011_svhn}
Yuval Netzer, Tao Wang, Adam Coates, Alessandro Bissacco, Bo Wu, and Andrew~Y. Ng.
\newblock Reading digits in natural images with unsupervised feature learning.
\newblock In {\em NIPS Workshop on Deep Learning and Unsupervised Feature Learning 2011}, 2011.

\bibitem{fedsim}
Chamath Palihawadana, Nirmalie Wiratunga, Anjana Wijekoon, and Harsha Kalutarage.
\newblock Fedsim: Similarity guided model aggregation for federated learning.
\newblock {\em Neurocomput.}, 483(C):432–445, apr 2022.

\bibitem{fada}
Xingchao Peng, Zijun Huang, Yizhe Zhu, and Kate Saenko.
\newblock Federated adversarial domain adaptation.
\newblock In {\em International Conference on Learning Representations}, 2020.

\bibitem{cpga}
Zhen Qiu, Yifan Zhang, Hongbin Lin, Shuaicheng Niu, Yanxia Liu, Qing Du, and Mingkui Tan.
\newblock Source-free domain adaptation via avatar prototype generation and adaptation.
\newblock In Zhi-Hua Zhou, editor, {\em Proceedings of the Thirtieth International Joint Conference on Artificial Intelligence, {IJCAI-21}}, pages 2921--2927. International Joint Conferences on Artificial Intelligence Organization, 8 2021.
\newblock Main Track.

\bibitem{bmd}
Sanqing Qu, Guang Chen, Jing Zhang, Zhijun Li, Wei He, and Dacheng Tao.
\newblock {BMD}: A general class-balanced multicentric dynamic prototype strategy for source-free domain adaptation.
\newblock In Shai Avidan, Gabriel Brostow, Moustapha Ciss{\'e}, Giovanni~Maria Farinella, and Tal Hassner, editors, {\em Computer Vision -- ECCV 2022}, pages 165--182, Cham, 2022. Springer Nature Switzerland.

\bibitem{snd}
Kuniaki Saito, Donghyun Kim, Piotr Teterwak, Stan Sclaroff, Trevor Darrell, and Kate Saenko.
\newblock Tune it the right way: Unsupervised validation of domain adaptation via soft neighborhood density.
\newblock In {\em 2021 IEEE/CVF International Conference on Computer Vision (ICCV)}, pages 9164--9173, 2021.

\bibitem{finch}
Saquib Sarfraz, Vivek Sharma, and Rainer Stiefelhagen.
\newblock Efficient parameter-free clustering using first neighbor relations.
\newblock In {\em 2019 IEEE/CVF Conference on Computer Vision and Pattern Recognition (CVPR)}, pages 8926--8935, 2019.

\bibitem{cfl}
Felix Sattler, Klaus-Robert Müller, and Wojciech Samek.
\newblock Clustered federated learning: Model-agnostic distributed multitask optimization under privacy constraints.
\newblock {\em IEEE Transactions on Neural Networks and Learning Systems}, 32(8):3710--3722, 2021.

\bibitem{som}
Mykola Servetnyk, Carrson~C. Fung, and Zhu Han.
\newblock Unsupervised federated learning for unbalanced data.
\newblock In {\em GLOBECOM 2020 - 2020 IEEE Global Communications Conference}, pages 1--6, 2020.

\bibitem{shenaj_wacv2023_ladd}
Donald Shenaj, Eros Fanì, Marco Toldo, Debora Caldarola, Antonio Tavera, Umberto Michieli, Marco Ciccone, Pietro Zanuttigh, and Barbara Caputo.
\newblock Learning across domains and devices: Style-driven source-free domain adaptation in clustered federated learning.
\newblock In {\em 2023 IEEE/CVF Winter Conference on Applications of Computer Vision (WACV)}, pages 444--454, 2023.

\bibitem{dropout}
Nitish Srivastava, Geoffrey Hinton, Alex Krizhevsky, Ilya Sutskever, and Ruslan Salakhutdinov.
\newblock Dropout: A simple way to prevent neural networks from overfitting.
\newblock {\em Journal of Machine Learning Research}, 15(56):1929--1958, 2014.

\bibitem{partialfed}
Benyuan Sun, Hongxing Huo, YI YANG, and Bo Bai.
\newblock Partialfed: Cross-domain personalized federated learning via partial initialization.
\newblock In M. Ranzato, A. Beygelzimer, Y. Dauphin, P.S. Liang, and J.~Wortman Vaughan, editors, {\em Advances in Neural Information Processing Systems}, volume~34, pages 23309--23320. Curran Associates, Inc., 2021.

\bibitem{fedka}
Yuwei Sun, Ng Chong, and Hideya Ochiai.
\newblock Feature distribution matching for federated domain generalization.
\newblock In Emtiyaz Khan and Mehmet Gonen, editors, {\em Proceedings of The 14th Asian Conference on Machine Learning}, volume 189 of {\em Proceedings of Machine Learning Research}, pages 942--957. PMLR, 12--14 Dec 2023.

\bibitem{PFL}
Alysa~Ziying Tan, Han Yu, Lizhen Cui, and Qiang Yang.
\newblock Towards personalized federated learning.
\newblock {\em IEEE Transactions on Neural Networks and Learning Systems}, 34(12):9587--9603, 2023.

\bibitem{fedpcl}
Yue Tan, Guodong Long, Jie Ma, LU LIU, Tianyi Zhou, and Jing Jiang.
\newblock Federated learning from pre-trained models: A contrastive learning approach.
\newblock In S. Koyejo, S. Mohamed, A. Agarwal, D. Belgrave, K. Cho, and A. Oh, editors, {\em Advances in Neural Information Processing Systems}, volume~35, pages 19332--19344. Curran Associates, Inc., 2022.

\bibitem{vdm}
Jiayi Tian, Jing Zhang, Wen Li, and Dong Xu.
\newblock Vdm-da: Virtual domain modeling for source data-free domain adaptation.
\newblock {\em IEEE Transactions on Circuits and Systems for Video Technology}, 32(6):3749--3760, 2022.

\bibitem{wscc}
Pu Tian, Weixian Liao, Wei Yu, and Erik Blasch.
\newblock Wscc: A weight-similarity-based client clustering approach for non-iid federated learning.
\newblock {\em IEEE Internet of Things Journal}, 9(20):20243--20256, 2022.

\bibitem{office-home}
Hemanth Venkateswara, Jose Eusebio, Shayok Chakraborty, and Sethuraman Panchanathan.
\newblock Deep hashing network for unsupervised domain adaptation.
\newblock In {\em 2017 IEEE Conference on Computer Vision and Pattern Recognition (CVPR)}, pages 5385--5394, 2017.

\bibitem{dipe}
Fan Wang, Zhongyi Han, Yongshun Gong, and Yilong Yin.
\newblock Exploring domain-invariant parameters for source free domain adaptation.
\newblock In {\em 2022 IEEE/CVF Conference on Computer Vision and Pattern Recognition (CVPR)}, pages 7141--7150, 2022.

\bibitem{semifedseg}
Huisi Wu, Baiming Zhang, Cheng Chen, and Jing Qin.
\newblock Federated semi-supervised medical image segmentation via prototype-based pseudo-labeling and contrastive learning.
\newblock {\em IEEE Transactions on Medical Imaging}, 43(2):649--661, 2024.

\bibitem{NEURIPS2021_f5deaeea}
Shiqi Yang, yaxing wang, Joost van~de Weijer, Luis Herranz, and Shangling Jui.
\newblock Exploiting the intrinsic neighborhood structure for source-free domain adaptation.
\newblock In M. Ranzato, A. Beygelzimer, Y. Dauphin, P.S. Liang, and J.~Wortman Vaughan, editors, {\em Advances in Neural Information Processing Systems}, volume~34, pages 29393--29405. Curran Associates, Inc., 2021.

\bibitem{yang2022attracting}
Shiqi Yang, Yaxing Wang, Kai Wang, SHANGLING JUI, and Joost van~de weijer.
\newblock Attracting and dispersing: A simple approach for source-free domain adaptation.
\newblock In Alice~H. Oh, Alekh Agarwal, Danielle Belgrave, and Kyunghyun Cho, editors, {\em Advances in Neural Information Processing Systems}, 2022.

\bibitem{dualadapt}
Chun-Han Yao, Boqing Gong, Hang Qi, Yin Cui, Yukun Zhu, and Ming-Hsuan Yang.
\newblock Federated multi-target domain adaptation.
\newblock In {\em Proceedings of the IEEE/CVF Winter Conference on Applications of Computer Vision (WACV)}, pages 1424--1433, January 2022.

\bibitem{fedseq}
Riccardo Zaccone, Andrea Rizzardi, Debora Caldarola, Marco Ciccone, and Barbara Caputo.
\newblock Speeding up heterogeneous federated learning with sequentially trained superclients.
\newblock In {\em 2022 26th International Conference on Pattern Recognition (ICPR)}, pages 3376--3382, 2022.

\bibitem{furl}
Fengda Zhang, Kun Kuang, Long Chen, Zhaoyang You, Tao Shen, Jun Xiao, Yin Zhang, Chao Wu, Fei Wu, Yueting Zhuang, et~al.
\newblock Federated unsupervised representation learning.
\newblock {\em Frontiers of Information Technology \& Electronic Engineering}, 24(8):1181--1193, 2023.

\bibitem{fedfomo}
Michael Zhang, Karan Sapra, Sanja Fidler, Serena Yeung, and Jose~M. Alvarez.
\newblock Personalized federated learning with first order model optimization.
\newblock In {\em International Conference on Learning Representations}, 2021.

\bibitem{Zhang_2023_CVPR}
Yixin Zhang, Zilei Wang, and Weinan He.
\newblock Class relationship embedded learning for source-free unsupervised domain adaptation.
\newblock In {\em 2023 IEEE/CVF Conference on Computer Vision and Pattern Recognition (CVPR)}, pages 7619--7629, 2023.

\bibitem{NEURIPS2022_215aeb07}
Ziyi Zhang, Weikai Chen, Hui Cheng, Zhen Li, Siyuan Li, Liang Lin, and Guanbin Li.
\newblock Divide and contrast: Source-free domain adaptation via adaptive contrastive learning.
\newblock In S. Koyejo, S. Mohamed, A. Agarwal, D. Belgrave, K. Cho, and A. Oh, editors, {\em Advances in Neural Information Processing Systems}, volume~35, pages 5137--5149. Curran Associates, Inc., 2022.

\bibitem{crl}
Zhengming Zhang, Yaoqing Yang, Zhewei Yao, Yujun Yan, Joseph~E. Gonzalez, Kannan Ramchandran, and Michael~W. Mahoney.
\newblock Improving semi-supervised federated learning by reducing the gradient diversity of models.
\newblock In {\em 2021 IEEE International Conference on Big Data (Big Data)}, pages 1214--1225, 2021.

\end{thebibliography}
}

\clearpage
\appendix
\section*{Appendix}

\section{Notations}
\label{app:notation}
\blue{The notations used in this paper are listed in \cref{table:notation}.}

\begin{table*}[t]
    \caption{\textbf{\blue{Notation in the paper.}} 
    }
    \centering
    \small
    \blue{
    \begin{tabular}{ll}\hline
        Symbol/Notation
        & Definition  \\ 
        \hline
        \hline
        \textbf{Problem} \\
        \hline
        $\mathcal{X}$ & feature space \\
        $\mathcal{Y}$ & label space \\
        $M$ & number of classes indexed by $m$\\
        $K$ & number of clients indexd by $k$ \\
        $R$ & number of communication rounds indexed by $r$ \\
        $E$ & number of local epochs for each client \\
        $D_k=\{x_{i}\}_{i=1}^{N_k}$ & unlabeled dataset of client $k$ with $N_k$ samples \\
        $\mathcal{P}_k(\mathcal{X})$ & data distribution of client $k$ \\
        $f \colon \mathcal{X} \rightarrow \mathbb{R}^q$ & feature extractor \\
        $g \colon \mathbb{R}^q \rightarrow \mathbb{R}^{M}$ & classifier \\
        $h = g \circ f$ & classification model \\
        $h_S = g_S \circ f_S$ & source model \\
        $W=[w_1,\dots,w_M]\in\mathbb{R}^{q\times M}$ & classifier weight for $g_S$ \\
        $h_k = g_k \circ f_k$ & locally trained model by client $k$ \\
        \hline
        \hline
        \textbf{Method} \\
        \hline
        $A(k,l)$ & adjacency matrix for client clustering \\
        $\kappa_k$ & nearest neighbor client for client $k$ \\
        $C$ & number of clusters indexed by $c$ \\
        $c_k$ & cluster index assigned to client $k$ \\
        $f_c$ & cluster model for cluster $c$ \\
        $\tilde{f}_c$ & soft cluster model for cluster $c$: $\tilde{f}_c = B_{c,0}f_c + B_{c,1}\sum_{c'}A_{c'\rightarrow c}f_{c'}$ \\
        $\bar{f}_k$ & locally combined model by client $k$ using soft cluster models: $\bar{f}_k=\sum_{c}\alpha_{k,c}\tilde{f}_c$ \\
        $f_k^{\rm init}$ & initial model of client $k$ for each round: $f_k^{\rm init} = \beta_{k,0} f_{c_k} + \beta_{k,1}\sum_{c}\alpha_{k,c}\tilde{f}_c$ \\
        $m(i)$ & class whose classifier weight vector is closest to $f_c(x_i)$: $m(i) = \argmax_m \cos(f_c(x_i), w_m)$ \\
        $\bm{v}_k=(v_{k,1},\dots,v_{k,C})$ & cluster weights for cluster models: $f_k^{\rm init} = \sum_{c}v_{k,c}f_c$ \\
        $\bm{\alpha}_k=(\alpha_{k,1},\dots,\alpha_{k,C})$ & locally calculated cluster weights for soft cluster models \\
        $\bm{\beta}_k = (\beta_{k,0}, \beta_{k,1})$ & pseudo-performance weights for $f_{c_k}$ and $\bar{f}_k$ \\
        $A_{c'\rightarrow c}$ & benefit metric indicating the relative advantage of cluster $c$ for clients in cluster $c'$ \\
        $B_{c,0}, B_{c,1}$ & coefficients balancing the emphasis between $f_c$ and $\sum_{c'}A_{c'\rightarrow c}f_{c'}$\\
        $T_a, T_b$ & temperature parameters controlling $\bm{\alpha}_k$ and $\bm{\beta}_k$, respectively\\
        $\lambda, \mu$ & balancing parameters for the loss function and mixup, respectively \\ 
        $\hat{D}_k=\{x_i,\hat{y}_i\}_{i=1}^{N_k}$ & pseudo-labeled dataset of client $k$ obtained by a prototype-based pseudo-labeling and mixup \\
        \hline
    \end{tabular}
    }
    \label{table:notation}
\end{table*}

\section{Details of Method}
\subsection{Algorithm}
\label{app:pseudo_code}
Algorithm~\ref{alg:fedwca} outlines the comprehensive training procedure for our suggested FedWCA, while Algorithm~\ref{alg:local_adaptation} details the local adaptation process.

\subsection{Soft Neighborhood Density} 
\label{app:SND}
We review Soft Neighborhood Density (SND) \cite{snd} used to assess the overall model efficacy in our FedWCA (see \cref{snd_weight}). SND is originally proposed as a method for evaluating models in an unsupervised manner by analyzing how densely data points cluster together using the overall models' outputs. It defines `soft neighborhoods' of a data point by the distribution of its similarity to other points, and measures density as the entropy of this distribution.

Let $h=g \circ f$ and $D=\{x_i\}_{i=1}^{N}$ be an evaluated model and an unlabeled dataset, respectively.
SND first computes the similarity between samples. Let $Q_{i,j}=\cos(h(x_i), h(x_j))$ be the $(i, j)$ element of the similarity matrix, where $\cos(\cdot, \cdot)$ is the cosine similarity. The diagonal elements of $Q$ are ignored to compute the distance to neighbors for each sample. $Q$ is then converted into a probabilistic distribution $P$ using the scaling temperature parameter $T$ and the softmax function:
\begin{align}
    P_{i,j} = \frac{\exp(Q_{i,j}/T)}{\sum_{j'}\exp(Q_{i,j'}/T)}.
\end{align}
The temperature is set to $0.05$ in the original paper and we use it. We finally obtain the SND value of $h$ by computing the entropy for each row of $P$ (\ie, each sample) and taking the average of all samples:
\begin{align}
    S(h) = - \frac{1}{N}\sum_{i=1}^{N}\sum_{j=1}^{N}P_{i,j}\log P_{i,j}.
\end{align}
See the original paper for detailed explanation.

\subsection{\blue{Cluster Weights}}
\label{app:specific_cluster_weights}
\blue{We provide the final specific cluster weights $\bm{v}_k = (v_{k,1},\dots,v_{k,C})$ in the initial model $f_k^{\rm init} = \sum_{c}v_{k,c}f_c$. 
According to \cref{aggregation_method}, the initial model for client $k$ is computed as follows:
\begin{align}
\label{weighted_init_model}
    &f_k^{\rm init} \\ \notag
    &= \beta_{k,0} f_{c_k} + \beta_{k,1}\sum_{c}\alpha_{k,c}\tilde{f}_c \\ \notag
    &= \beta_{k,0} f_{c_k} + \beta_{k,1}\sum_{c}\alpha_{k,c}\{B_{c,0}f_c + B_{c,1}\sum_{c'}A_{c'\rightarrow c}f_{c'}\} \\ \notag
    &= \sum_c \{\mathbbm{1}(c=c_k)\beta_{k,0} + \beta_{k,1}\alpha_{k,c}B_{c,0} \\ \notag 
    & \qquad \qquad \qquad \qquad \qquad+\beta_{k,1}\sum_{c'}B_{c',1}\alpha_{k,c'}A_{c\rightarrow c'}\}f_c,
\end{align}
where $\mathbbm{1}$ is the indicator function. The third equation can be obtained by replacing the subscripts $c$ and $c'$. We thus obtain the cluster weights $\bm{v}_k$ as:
\begin{align}
    v_{k,c} = \mathbbm{1}(c=c_k)\beta_{k,0} &+ \beta_{k,1}\alpha_{k,c}B_{c,0} \\ \notag 
    & +\beta_{k,1}\sum_{c'}B_{c',1}\alpha_{k,c'}A_{c\rightarrow c'}.
\end{align}
}

\subsection{Loss Function of SHOT}
\label{app:loss}
We review the loss function of SHOT \cite{liang_icml2020_shot}, adopted as a loss function of our method because of its simplicity. 
The loss function comprises the cross-entropy loss for the pseudo-labeled dataset and an Information Maximization (IM) loss \cite{im_loss} for the unlabeled dataset. 
The IM loss promotes confident model outputs and counteracts a bias towards any single class by discouraging trivial score distributions. The total loss function $\mathcal{L}(h)$ for the model $h$ weights these two loss functions with a balancing parameter $\lambda$: 
\begin{align}
    \label{loss_function}
    &\mathcal{L}(h) = \mathcal{L}_{\rm IM}(h;D) + \lambda \mathcal{L}_{\rm CE}(h;\hat{D}); \\ \notag
    &\mathcal{L}_{\rm CE}(h;\hat{D}) = - \frac{1}{N}\sum_{(x,\hat{y})\in \hat{D}}\log{(h(x))_{\hat{y}}}, \\ \notag
    &\mathcal{L}_{\rm IM}(h;D) = \sum_{m=1}^M\hat{o}_m\log{\hat{o}_m} \\ \notag
    &\qquad \qquad \qquad - \frac{1}{N}\sum_{m=1}^M\sum_{x\in D}(h(x))_m\log{(h(x))_m},
\end{align}
where $N$ is the dataset size and $\mathcal{L}_{\rm CE}(h;\hat{D})$ represents the cross-entropy loss calculated over the pseudo-labeled dataset $\hat{D}$, which encourages the model $h$ to align with the pseudo-labels.
The IM loss $\mathcal{L}_{\rm IM}(h;D)$, on the other hand, comprises two terms: (1) the negative entropy of the mean output score $\hat{o}=\sum_{x \in D}h(x)/N$ and (2) the entropy of the model's output scores for each data point in the unlabeled dataset $D$.

\section{Detailed Experimental Settings}
\label{app:experimental_details}
\subsection{Datasets}
\label{app:dataset}
\cref{table:dataset_information} summarizes the details of datasets used in our experiments. For Digit-Five, we used the entire dataset for USPS and 34,000 randomly selected samples for the other domains. For each dataset, one domain was selected as the source, and the remaining domains served as target domains. Data samples from each target domain were distributed to 8 clients for Digit-Five and 3 clients for PACS and Office-Home per domain, all in an i.i.d. manner. Each client's data were divided into three subsets: 20\% for testing, 64\% for training, and 16\% for validation.

\begin{table*}[t]%
    \caption{\textbf{Dataset information} including number of samples, classes, and clients for three datasets.}
    \centering
    \begin{tabular}{rcccc}\hline
        \multirow{2}{*}{\textbf{Datasets}} &  \multicolumn{2}{c} {\textbf{Number of samples}} & \multirow{2}{*}{\textbf{Classes}} & \multirow{2}{*}{\textbf{Clients}} \\
        \cline{2-3}
        & Per domain & Total &  &  \\ \hline
        \multirow{2}{*}{Digit-Five} &  MNIST: 34,000, MNIST-M: 34,000 & \multirow{2}{*}{145,298} & \multirow{2}{*}{10} & \multirow{2}{*}{32}\\
        & SVHN: 34,000, SYNTH: 34,000, USPS: 9298 & & & \\
        \hline
        \multirow{2}{*}{PACS} & Art Painting: 2,048, Cartoon: 2,344 & \multirow{2}{*}{9,991} & \multirow{2}{*}{7} & \multirow{2}{*}{9}\\
        & Photo: 1,670, Sketch: 3,929 & & & \\
        \hline
        \multirow{2}{*}{Office-Home} & Art: 2,424, Clipart: 4,365 & \multirow{2}{*}{15,579} & \multirow{2}{*}{65} & \multirow{2}{*}{9}\\
        & Product: 4,437, Real-World: 4,353 & & & \\
        \hline
    \end{tabular}
    \label{table:dataset_information}
\end{table*}

\begin{table}[t]
    \caption{\textbf{Hyperparameters for our FedWCA} including the learning rate (lr), balancing parameter for cross-entropy loss $\lambda$, balancing parameter for mixup $\mu$, and temperature parameters $T_a$, $T_b$. Only for SYNTH of Digit-Five and Clipart of Office-Home source domains, $T_a$ is set to 0.001 and 0.1, respectively.}
    \centering
    \begin{tabular}{rccccc}\hline
        \multirow{2}{*}{\textbf{Datasets}} &  \multicolumn{5}{c} {\textbf{Hyperparameters}} \\
        \cline{2-6}
        & lr & $\lambda$ & $\mu$ & $T_a$ & $T_b$ \\ \hline
        Digit-Five & 0.001 & \ 0.1 \ & 0.55 & 0.01 & 0.05 \\
        PACS & 0.0001 & 0.3 & 0.55 & 0.1 & 0.05 \\
        Office-Home & 0.0001 & 0.3 & 0.55 & 0.01 & 0.05 \\
        \hline
    \end{tabular}
    \label{table:hyparams}
\end{table}

\begin{figure*}[tb]%
  \centering
  \begin{subfigure}{0.48\linewidth}
    \centering
    \includegraphics[width=0.8\linewidth]{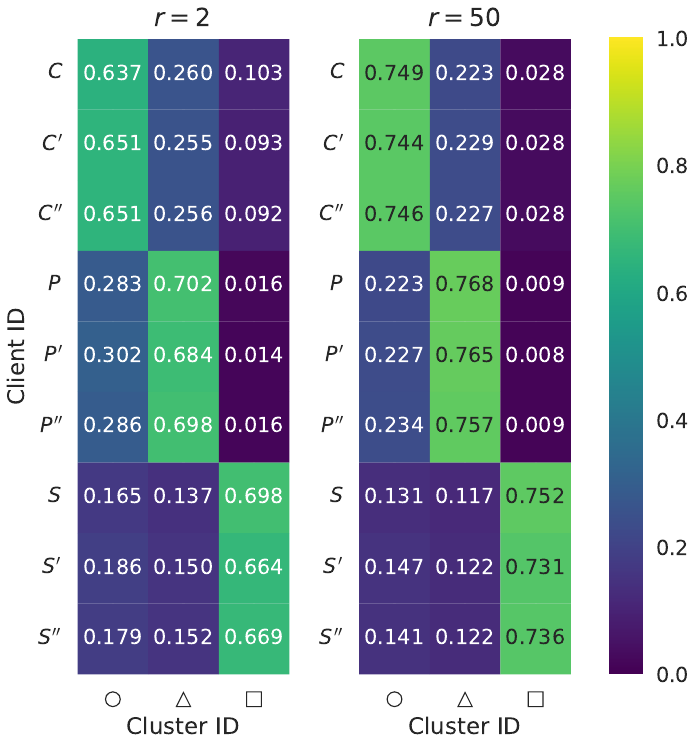}
    \caption{FedWCA-L}
  \end{subfigure}
  \hfill
  \begin{subfigure}{0.48\linewidth}
    \centering
    \includegraphics[width=0.8\linewidth]{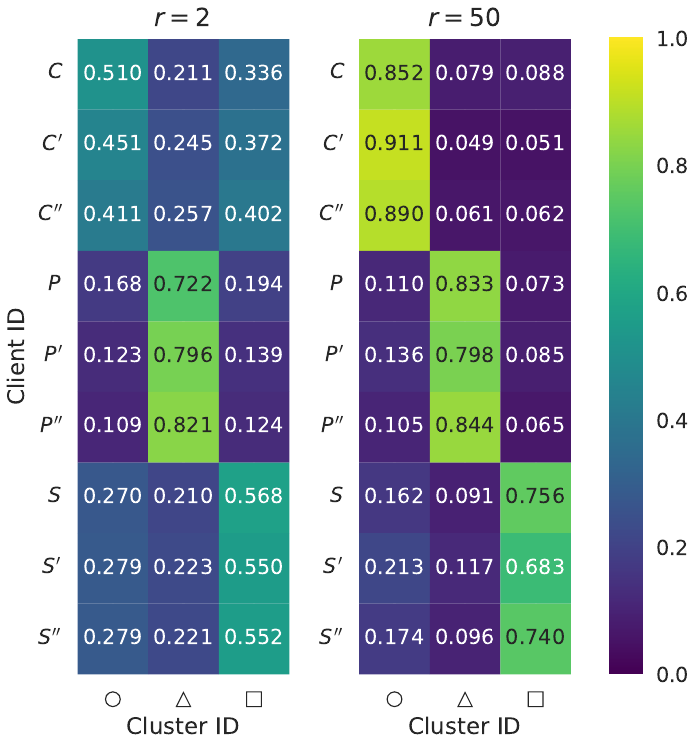}
    \caption{FedWCA}
  \end{subfigure}
  \caption{\textbf{Visualization of cluster weights} for FedWCA-L and FedWCA when the communication round $r$ is $2$ and $50$. Art Painting in PACS dataset is used for the source dataset. 
  Clients $C, C', C''$ belong to Cartoon, $P, P', P''$ to Photo, and $S, S', S''$ to Sketch. See the first row of \cref{table:clustering_result} for the cluster IDs (denoted by $\bigcirc$, $\triangle$, and $\square$) assigned to each client. The number represents the weight for the corresponding cluster.
  }
  \label{fig:cluster_weights}
\end{figure*}

\subsection{Implementation Details}
\label{app:implementation}

For the implementation, we set the local epoch to $E=5$ and total communication rounds to $R=100$ for Digit-Five and PACS, and $R=50$ for Office-Home, ensuring convergence of each method's learning. 
In the case of FedPCL+PL, we limited $R$ to 20, due to overfitting observed in this method on the PACS and Office-Home datasets. The stochastic gradient descent (SGD) was used with \blue{the best} learning rate of $10^{-3}$ for Digit-Five and $10^{-4}$ for PACS and Office-Home in all methods, \blue{which are selected from $\{10^{-i}|i\in\{1,2,3,4,5\}\}$}. For source model training, we set the learning rate to $0.001$ in all datasets.  Additionally, we adopted a weight decay of $0.001$ and a momentum of $0.9$, in line with standard SFDA studies \cite{liang_icml2020_shot}. In LADD's implementation, we randomly selected 100 samples from each client's target data for computing style features, focusing on the central 5\% region of the frequency spectrum. Other hyperparameters for LADD were also appropriately tuned. The temperature parameters of FedPCL were set to $70$ as a result of tuning. For our methods, the temperature parameters were set to $T_b=0.05$ for all datasets (aligned with SND \cite{snd}) and $T_a=0.01$ for Digit-Five and Office-Home, except for SYNTH ($T_a=0.001$) and Clipart ($T_a=0.1$) source domains, while $T_a=0.1$ for PACS. The balancing parameter $\lambda$ was assigned the same value as in the original SHOT paper\cite{liang_icml2020_shot}. In particular, \cref{table:hyparams} summarizes the hyperparameters for FedWCA.

\subsection{\blue{Hyperparameter Tuning}}
\blue{For LADD, we searched the regularization parameters and starting rounds from $\{10^{-i}|i\in\{-2, -1, 0, 1,2,3,4,5\}\}$ and $\{5, 10, 30, 50, 80, 100\}$, respectively. For FedPCL, the best temperature parameter was selected from $\{7 \times 10^{-i}|i\in\{-2,-1,0, 1, 2, 3\}\}$ as par the original paper. For our FedWCA, we searched the temperature parameters $T_a$ and $T_b$ from $\{0.001, 0.005, 0.01, 0.05, 0.1\}$.}

\section{Limitations and Discussions}
\label{app:limitation}

\textbf{Complexity of FedWCA.}
We mention that the derivation of cluster weights in our method, while not computationally intensive, involves a complex procedure. Although \cref{sec:ablation_study} demonstrates that simpler weight calculations do not provide sufficient performance, identifying simpler yet effective methods for calculating cluster weights that match FedWCA's performance remains a task for future research. However, our method can be slightly modified to reduce computational costs, as discussed below.

\textbf{Concerns about distributing all cluster models.}
Our method requires each client to receive all soft cluster models, which may raise some concerns: (1) privacy concerns and (2) additional communication and storage costs. 

\begin{table}[t]
    \caption{\textbf{Results for FedWCA modification reducing costs.} The numbers are the mean values $\pm$ standard deviations of the averaged accuracy (\%) across all clients and all target domains. In the latter $U-1$ rounds of every $U$ round, the initial model of the FedWCA can be computed server-side as opposed to client-side, thereby cutting communication, storage, and computational costs. This revised FedWCA boasts an equivalent performance to its original counterpart when $U=5$.} 
    \label{tab:modified_fedwca}
    \centering
    \footnotesize
    \begin{tabular}{cccc}\hline
        \multirow{3}{*}{\textbf{Datasets}} &  & \multicolumn{2}{c} {\textbf{Methods}} \\
        \cline{3-4}
        & & Revised FedWCA & \multirow{2}{*}{FedWCA} \\
        & & $U=5$ & \\
        \hline
        \multirow{6}{*}{\textbf{Digit-Five}} & MN & 73.37 $\pm$ 3.65 & 72.74 $\pm$ 3.57 \\
        & SV & 89.88 $\pm$ 2.24 & 90.13 $\pm$ 2.08 \\
        & MN-M & 83.16 $\pm$ 0.56 & 82.93 $\pm$ 0.79 \\
        & US & 57.33 $\pm$ 3.00 & 58.56 $\pm$ 4.26 \\
        & SY & 85.70 $\pm$ 1.05 & 84.06 $\pm$ 1.38 \\
        \cline{2-4}
        & \textbf{Avg.} & 77.89 $\pm$ 11.93 & 77.69 $\pm$ 11.47 \\
        \hline 
        \multirow{5}{*}{\textbf{PACS}} & Ar & 80.89 $\pm$ 2.54 & 80.63 $\pm$ 2.44 \\
        & Ca & 83.23 $\pm$ 0.91 & 83.18 $\pm$ 0.86 \\
        & Ph & 65.32 $\pm$ 1.06 & 65.50 $\pm$ 1.24 \\
        & Sk & 84.28 $\pm$ 6.40 & 84.22 $\pm$ 6.60 \\
        \cline{2-4}
        & \textbf{Avg.} & 78.43 $\pm$ 8.45 & 78.38 $\pm$ 8.38 \\
        \hline
        \multirow{5}{*}{\textbf{Office-Home}} & Ar & 66.13 $\pm$ 0.67 & 66.06 $\pm$ 0.50 \\
        & Cl & 68.33 $\pm$ 0.61 & 68.32 $\pm$ 0.42 \\
        & Pr & 61.30 $\pm$ 0.60 & 61.46 $\pm$ 0.88 \\
        & Re & 67.63 $\pm$ 0.29 & 68.06 $\pm$ 0.46 \\
        \cline{2-4}
        & \textbf{Avg.} & 65.85 $\pm$ 2.86 & 65.97 $\pm$ 2.83 \\
        \hline
    \end{tabular}
    \label{table:full_pacs}
\end{table}

(1) Distributing cluster models poses minimal privacy risks due to several factors. Initially, we note that the server distributes all "soft cluster models" to each client, along with the original cluster model of the respective client. Soft cluster models, produced by merging all cluster models, roughly incorporate all clients' local models and thus, privacy risks remain similar to typical FL methods such as FedAvg. Further, distributing the single original cluster model to clients is a standard practice in current clustered FL studies and carries minimal privacy risk. This is due to our algorithm's design, where clients are only privy to their specific cluster IDs with no insight into other cluster details such as cluster size or client membership, even if it is composed of only two clients.


(2) 
Our weighted cluster aggregation (WCA) costs $C+1$ times higher than FedAvg ($C$: number of clusters) due to the requirement of each client receiving and storing $C$ soft cluster models and the original cluster model. However, we can save those costs by altering the computation of the client's initial models in WCA as follows. In the first round of every $U$ rounds, clients compute their initial models as the original method does. In the following $U-1$ rounds, the initial models are computed server-side, utilizing identical cluster weights sent by clients in the first round. This modification allows the server to only distribute two models to each client: the computed initial model and the original cluster model, reducing the communication and storage costs by a factor of $2/(C+1)$ in $(U-1)/U$ of all rounds. This can also lessen the client's computational cost for initial model calculations. As demonstrated in \cref{tab:modified_fedwca}, further tests on Digit-Five, PACS, and Office-Home with $U=5$ showed that the revised FedWCA maintains similar average accuracy as the original. Notably, in some source domains like SYNTH, the revised technique improves accuracy. This enhancement is because the use of identical cluster weights reduces overfitting.

%



\textbf{Different enhancement according to dataset.}
As illustrated in \cref{sec:results}, the accuracy enhancement achieved by our method varies with the dataset. Specifically, for datasets exhibiting minor domain gaps, FedAvg may suffice to some extent, as the impact of domain shifts on accuracy is minimal. For instance, in the Office-Home dataset, although our method surpasses FedAvg, the superiority margin is less significant than in Digit-Five and PACS.

\textbf{Extension to other tasks.}
While our approach is primarily tailored for classification tasks, its underlying principles could theoretically extend to other complex vision tasks, like object detection. However, direct extrapolation may encounter challenges, such as convergence of feature vectors towards classifier vectors in cluster weight calculations.
Adapting and assessing our method in these diverse contexts constitutes an avenue for future research.

\section{Supplementary Experimental Results}

\begin{table}[tb]
    \centering
    \caption{\textbf{Full results for layer dependence of clustering.} Art Painting of PACS is used as the source domain, and ResNet-18 is employed. Clients $C, C', C''$ belong to Cartoon, $P, P', P''$ to Photo, and $S, S', S''$ to Sketch. The table below reports the cluster IDs (denoted by $\bigcirc$, $\triangle$, and $\square$) assigned to each client. 
    Ideally, each client group $(C, C', C'')$, $(P, P', P'')$, and $(S, S', S'')$ should be grouped together, as is achieved by the first and second layers.
    }

    \label{table:clustering_full_result}
    \footnotesize
    \begin{tabular}{rccccccccc}\hline
        \textbf{ } &  \multicolumn{9}{c} {\textbf{Clients}} \\
        \cline{2-10}
        \textbf{Layer} & $C$ & $C'$ & $C''$ & $P$ & $P'$ & $P''$ & $S$ & $S'$ & $S''$ \\ \hline
        1st & $\bigcirc$ & $\bigcirc$ & $\bigcirc$ & $\triangle$ & $\triangle$ & $\triangle$ & $\square$ & $\square$ & $\square$ \\
        2nd & $\bigcirc$ & $\bigcirc$ & $\bigcirc$ & $\triangle$ & $\triangle$ & $\triangle$ & $\square$ & $\square$ & $\square$ \\
        3rd & $\bigcirc$ & $\bigcirc$ & $\bigcirc$ & $\bigcirc$ & $\bigcirc$ & $\bigcirc$ & $\bigcirc$ & $\bigcirc$ & $\bigcirc$ \\
        4th & $\bigcirc$ & $\bigcirc$ & $\bigcirc$ & $\bigcirc$ & $\bigcirc$ & $\bigcirc$ & $\triangle$ & $\triangle$ & $\triangle$ \\
        5th & $\bigcirc$ & $\bigcirc$ & $\bigcirc$ & $\bigcirc$ & $\bigcirc$ & $\bigcirc$ & $\bigcirc$ & $\bigcirc$ & $\bigcirc$ \\
        6th & $\bigcirc$ & $\bigcirc$ & $\bigcirc$ & $\triangle$ & $\triangle$ & $\triangle$ & $\square$ & $\square$ & $\square$ \\
        7th & $\bigcirc$ & $\bigcirc$ & $\bigcirc$ & $\bigcirc$ & $\bigcirc$ & $\bigcirc$ & $\bigcirc$ & $\bigcirc$ & $\bigcirc$ \\
        8th & $\bigcirc$ & $\bigcirc$ & $\bigcirc$ & $\bigcirc$ & $\bigcirc$ & $\bigcirc$ & $\bigcirc$ & $\bigcirc$ & $\bigcirc$ \\
        9th & $\bigcirc$ & $\bigcirc$ & $\bigcirc$ & $\bigcirc$ & $\bigcirc$ & $\bigcirc$ & $\bigcirc$ & $\bigcirc$ & $\bigcirc$ \\
        10th & $\bigcirc$ & $\bigcirc$ & $\bigcirc$ & $\bigcirc$ & $\bigcirc$ & $\bigcirc$ & $\triangle$ & $\triangle$ & $\triangle$ \\
        11th & $\bigcirc$ & $\bigcirc$ & $\bigcirc$ & $\bigcirc$ & $\bigcirc$ & $\bigcirc$ & $\bigcirc$ & $\bigcirc$ & $\bigcirc$ \\
        12th & $\bigcirc$ & $\bigcirc$ & $\bigcirc$ & $\bigcirc$ & $\bigcirc$ & $\bigcirc$ & $\bigcirc$ & $\triangle$ & $\triangle$ \\
        13th & $\bigcirc$ & $\bigcirc$ & $\bigcirc$ & $\bigcirc$ & $\bigcirc$ & $\bigcirc$ & $\bigcirc$ & $\bigcirc$ & $\bigcirc$ \\
        14th & $\bigcirc$ & $\bigcirc$ & $\bigcirc$ & $\triangle$ & $\triangle$ & $\triangle$ & $\square$ & $\square$ & $\square$ \\
        15th & $\bigcirc$ & $\bigcirc$ & $\bigcirc$ & $\bigcirc$ & $\bigcirc$ & $\bigcirc$ & $\bigcirc$ & $\bigcirc$ & $\bigcirc$ \\
        16th & $\bigcirc$ & $\bigcirc$ & $\bigcirc$ & $\bigcirc$ & $\bigcirc$ & $\bigcirc$ & $\triangle$ & $\triangle$ & $\triangle$ \\
        17th & $\bigcirc$ & $\bigcirc$ & $\bigcirc$ & $\bigcirc$ & $\bigcirc$ & $\bigcirc$ & $\bigcirc$ & $\bigcirc$ & $\bigcirc$ \\
        All & $\bigcirc$ & $\bigcirc$ & $\bigcirc$ & $\bigcirc$ & $\bigcirc$ & $\bigcirc$ & $\bigcirc$ & $\bigcirc$ & $\bigcirc$ \\
        \hline
    \end{tabular}
\end{table}

\subsection{Visualization of Cluster Weights}
\label{app:cluster_weights}
\cref{fig:cluster_weights} visualizes the cluster weights for our proposed methods FedWCA and FedWCA-L, a variant of FedWCA wherein cluster weights are computed solely locally (see \cref{aggregation_method}). This shows that clients within the same domain obtain similar weights, indicating the efficacy of our cluster weight calculation based on \cref{cluster_weights}. FedWCA, in particular, promotes inter-domain collaboration early in learning ($r=2$), notably between Cartoon and Sketch domains. Specifically, Cartoon domain clients ($C, C', C''$) have significantly higher weights for cluster $\square$ (Sketch) in FedWCA compared to FedWCA-L. This is the result of the Cartoon domain clients considering the benefits for the clients in cluster $\square$. As learning progresses ($r=50$), FedWCA clients increasingly concentrate on their respective clusters, highlighting the method's balance between overall and individual advantages. As a result, this approach leads to notable accuracy improvements for both Cartoon and Sketch clients, as depicted in \cref{fig:accuracy_plot} (b).


\begin{table}[t]
    \centering
    \caption{\textbf{Results of client clustering in Digit-Five} for each source domain. The clustering algorithm FINCH is applied to Lenet. Each row represents client IDs. Clients $1$ to $8$, $9$ to $16$, $17$ to $24$ and $25$ to $32$ each belong to the same domain. The table below reports the cluster IDs (denoted by, \eg, $\bigcirc$, $\triangle$, $\square$, and $\diamondsuit$) assigned to each client. 
    Ideally, each client group (1,2,3,4,5,6,7,8), (9,10,11,12,13,14,15,16), (17,18,19,20,21,22,23,24), and (25,26,27,28,29,30,31,32) should be grouped together, as is achieved by SVHN source domain.}
    \label{table:clustering_result_digit}
    \small
    \begin{tabular}{cccccc}\hline
        \multirow{2}{*}{\textbf{Client ID}} &  \multicolumn{5}{c} {\textbf{Source domains}} \\
        \cline{2-6}
        & \ \ MN \ \ & SV & MN-M & US & \ \ SY \ \ \\ \hline
        1 & $\bigcirc$ & $\bigcirc$ & $\bigcirc$ & $\bigcirc$ & $\bigcirc$ \\
        2 & $\bigcirc$ & $\bigcirc$ & $\bigcirc$ & $\bigcirc$ & $\bigcirc$ \\
        3 & $\bigcirc$ & $\bigcirc$ & $\bigcirc$ & $\bigcirc$ & $\bigcirc$ \\
        4 & $\bigcirc$ & $\bigcirc$ & $\bigcirc$ & $\bigcirc$ & $\bigcirc$ \\
        5 & $\bigcirc$ & $\bigcirc$ & $\bigcirc$ & $\bigcirc$ & $\bigcirc$ \\
        6 & $\bigcirc$ & $\bigcirc$ & $\bigcirc$ & $\bigcirc$ & $\bigcirc$ \\
        7 & $\triangle$ & $\bigcirc$ & $\bigcirc$ & $\bigcirc$ & $\bigcirc$ \\
        8 & $\triangle$ & $\bigcirc$ & $\bigcirc$ & $\bigcirc$ & $\bigcirc$ \\
        \hline
        9 & $\square$ & $\triangle$ & $\triangle$ & $\triangle$ & $\triangle$ \\
        10 & $\square$ & $\triangle$ & $\triangle$ & $\triangle$ & $\triangle$ \\
        11 & $\square$ & $\triangle$ & $\triangle$ & $\triangle$ & $\triangle$ \\
        12 & $\square$ & $\triangle$ & $\triangle$ & $\triangle$ & $\triangle$ \\
        13 & $\diamondsuit$ & $\triangle$ & $\triangle$ & $\triangle$ & $\triangle$ \\
        14 & $\diamondsuit$ & $\triangle$ & $\triangle$ & $\triangle$ & $\triangle$ \\
        15 & $\heartsuit$ & $\triangle$ & $\triangle$ & $\triangle$ & $\triangle$ \\
        16 & $\heartsuit$ & $\triangle$ & $\triangle$ & $\triangle$ & $\triangle$ \\
        \hline
        17 & $\bigstar$ & $\square$ & $\square$ & $\square$ & $\square$ \\
        18 & $\bigstar$ & $\square$ & $\square$ & $\square$ & $\square$ \\
        19 & $\bigstar$ & $\square$ & $\square$ & $\square$ & $\square$ \\
        20 & $\bigstar$ & $\square$ & $\square$ & $\square$ & $\square$ \\
        21 & $\bigstar$ & $\square$ & $\square$ & $\square$ & $\diamondsuit$ \\
        22 & $\bigstar$ & $\square$ & $\diamondsuit$ & $\diamondsuit$ & $\diamondsuit$ \\
        23 & $\clubsuit$ & $\square$ & $\diamondsuit$ & $\diamondsuit$ & $\diamondsuit$ \\
        24 & $\clubsuit$ & $\square$ & $\diamondsuit$ & $\diamondsuit$ & $\diamondsuit$ \\
        \hline
        25 & $\spadesuit$ & $\diamondsuit$ & $\heartsuit$ & $\heartsuit$ & $\triangle$ \\
        26 & $\spadesuit$ & $\diamondsuit$ & $\heartsuit$ & $\heartsuit$ & $\triangle$ \\
        27 & $\spadesuit$ & $\diamondsuit$ & $\heartsuit$ & $\heartsuit$ & $\triangle$ \\
        28 & $\spadesuit$ & $\diamondsuit$ & $\heartsuit$ & $\heartsuit$ & $\triangle$ \\
        29 & $\spadesuit$ & $\diamondsuit$ & $\heartsuit$ & $\heartsuit$ & $\triangle$ \\
        30 & $\spadesuit$ & $\diamondsuit$ & $\heartsuit$ & $\heartsuit$ & $\triangle$ \\
        31 & $\spadesuit$ & $\diamondsuit$ & $\heartsuit$ & $\heartsuit$ & $\triangle$ \\
        32 & $\spadesuit$ & $\diamondsuit$ & $\heartsuit$ & $\heartsuit$ & $\triangle$ \\
        \hline
    \end{tabular}
\end{table}

\begin{table}[t]
    \caption{\textbf{Ablation on clustering algorithm.} Three clustering algorithms are applied to FedWCA. `True' uses the hypothetical true clusters based on the domain labels. Even without specific cluster information, FedWCA offers similar performance to `True'.}
    \label{table:different_clustering}
    \centering
    \footnotesize
    \begin{tabular}{rcccccc}\hline
        \multirow{2}{*}{\textbf{Clustering}} &  \multicolumn{6}{c} {\textbf{Digit-Five}}\\
        \cline{2-7}
        & MN & SV & MN-M & US  & SY & \textbf{Avg.} \\ 
        \hline
        FINCH & 72.74 & 90.13 & 83.93 & 58.56 & 84.94 & 77.86 \\
        LADD & 72.33 & 89.89 & 81.99 & 58.72 & 85.58 & 77.70 \\
        True & 72.93 & 90.07 & 83.43 & 56.95 & 85.69 & 77.81 \\
        \hline
    \end{tabular}
\end{table}

\subsection{Results of Client Clustering}
\label{app:clustering}

\cref{table:clustering_full_result} shows the full results for the layer dependence of client clustering in our method with the PACS dataset.

\begin{figure*}[t]
    \centering
    \includegraphics[width=0.8\linewidth]{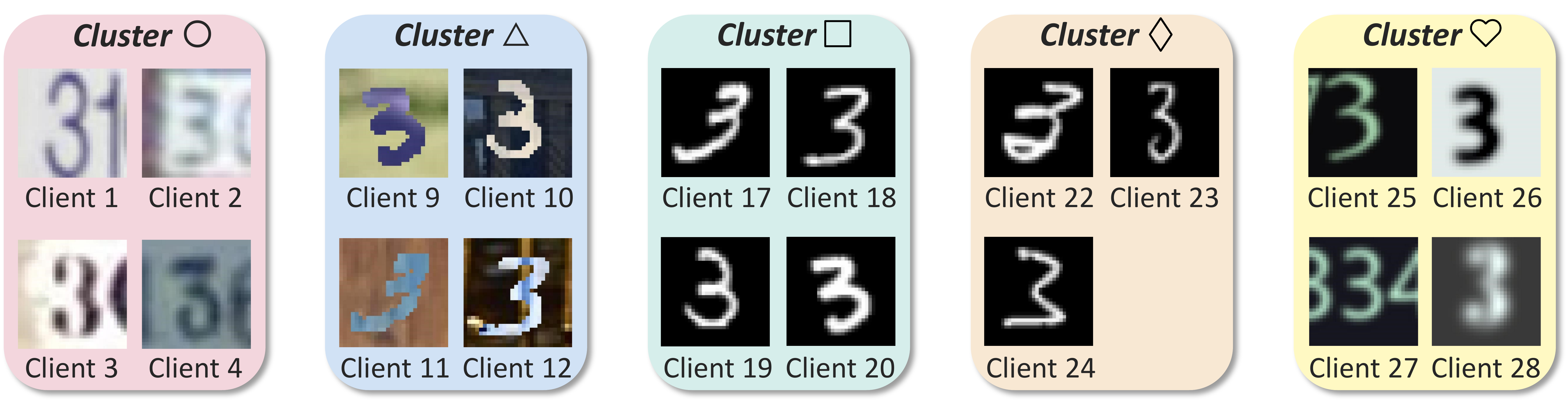}
    \caption{\blue{\textbf{Images possessed by clients in each cluster.} USPS of Digit-Five serves as the source domain. Our FedWCA clusters clients according to their local models, almost matching their data domains.}}
    \label{fig:images}
\end{figure*}

Additionally, \cref{table:clustering_result_digit} presents the clustering results for Digit-Five across various source domains using our FINCH method. \blue{\cref{fig:images} shows the specific images possessed by clients in each cluster particularly when using USPS as the source domain.} The results from \cref{table:clustering_result_digit} indicate that, unlike PACS and Office-Home, the clustering for Digit-Five contains some errors, except for the SVHN source domain.
To this end, we compare our standard FINCH algorithm, against two alternatives on Digit-Five: (1) LADD clustering based on the shared style features and (2) a hypothetical `true' clustering based on actual domain information. It must be emphasized that LADD requires a specified number of iterations and a search range for clusters, whereas FINCH necessitates no hyperparameters. \cref{table:different_clustering} shows the accuracy of FedWCA when applying each clustering algorithm. Notably, FINCH performs comparably to both the true clustering and LADD, with no need for sharing prior information on the clusters or any information other than the model parameters. This highlights our method's strength in enhancing client performance, even when assigned to incorrect clusters, by effectively combining all cluster models using individualized cluster weights.

\begin{table*}[th]
    \caption{\blue{\textbf{Performance comparison with extended FL methods (Digit-Five and PACS).} 
    The existing FL methods, FedProx \cite{fedprox}, FedAMP \cite{fedamp}, and IFCA \cite{ifca} are extended to FFREEDA by incorporating \textbf{PP}: prototype-based pseudo-labeling, \textbf{FP}: fixed pseudo-labels in each round, and \textbf{IM}: IM loss. FedWCA outperforms them in all source domains.
    }} 
    \label{tab:additional_comparison}
    \centering
    \small
    \blue{
    \begin{tabular}{cccccc}\hline
        \multirow{3}{*}{\textbf{Datasets}} &  & \multicolumn{4}{c} {\textbf{Methods} (+PP+FP+IM)} \\
        \cline{3-6}
        & & FedProx & FedAMP & IFCA & FedWCA (ours) \\
        \hline
        \multirow{6}{*}{\textbf{Digit-Five}} & MN & 56.01 $\pm$ 2.46 & 51.48 $\pm$ 0.44 & 58.68 $\pm$ 2.21 & \textbf{72.74} $\pm$ \textbf{3.57} \\
        & SV & 83.27 $\pm$ 1.80 & 83.00 $\pm$ 0.81 & 83.39 $\pm$ 1.60 & \textbf{90.13} $\pm$ \textbf{2.08} \\
        & MN-M & 81.27 $\pm$ 0.29 & 72.94 $\pm$ 2.09 & 79.82 $\pm$ 0.40 & \textbf{82.93} $\pm$ \textbf{0.79}  \\
        & US & 52.78 $\pm$ 0.74 & 53.81 $\pm$ 1.54 & 55.62 $\pm$ 2.99 & \textbf{58.56} $\pm$ \textbf{4.26} \\
        & SY & 82.21 $\pm$ 0.54 & 82.22 $\pm$ 0.32 & 81.71 $\pm$ 0.75 & \textbf{84.94} $\pm$ \textbf{1.15} \\
        \cline{2-6}
        & \textbf{Avg.} & 71.11 $\pm$ 13.86 & 68.69 $\pm$ 13.73 & 71.81 $\pm$ 12.30 & \textbf{77.86} $\pm$ \textbf{11.56} \\
        \hline 
        \multirow{5}{*}{\textbf{PACS}} & Ar & 76.50 $\pm$ 1.24 & 73.54 $\pm$ 1.30 & 75.82 $\pm$ 3.11 & \textbf{80.63} $\pm$ \textbf{2.44} \\
        & Ca & 76.16 $\pm$ 0.87 & 77.55 $\pm$ 1.21 & 79.13 $\pm$ 2.57 & \textbf{83.18} $\pm$ \textbf{0.86} \\
        & Ph & 60.88 $\pm$ 0.74 & 63.70 $\pm$ 0.79 & 64.61 $\pm$ 1.57 & \textbf{65.50} $\pm$ \textbf{1.24} \\
        & Sk & 80.29 $\pm$ 7.02 & 75.84 $\pm$ 5.87 & 80.93 $\pm$ 6.59 & \textbf{84.22 $\pm$ 6.60} \\
        \cline{2-6}
        & \textbf{Avg.} & 73.46 $\pm$ 8.29 & 72.66 $\pm$ 6.19 & 75.13 $\pm$ 7.46 & \textbf{78.38 $\pm$ 8.38} \\
        \hline
    \end{tabular}
    }
\end{table*}

\subsection{\blue{Comparison with Other Federated Learning Methods}}
\label{app:additional_comparison}

\blue{In addition to FedAvg and FedPCL, we extended other FL methods using ground-truth labels to FFREEDA by incorporating a prototype-based pseudo-labeling, fixed pseudo-labels per round, and IM loss, and compared them to our method. We adopted three additional FL methods: (1) FedProx \cite{fedprox}, (2) FedAMP \cite{fedamp}, and (3) IFCA \cite{ifca}. FedProx modifies FedAvg by adding the proximal term to address data heterogeneity. 
FedAMP personalizes FL by creating a client-specific cloud model weighted on the similarity of each client's model parameters, with clients subsequently training personalized models based on this cloud model.
IFCA, a clustered FL method that requires setting the number of clusters, has each client calculate the loss for all cluster models and train the model with the lowest loss per round.

The performance comparison is shown in \cref{tab:additional_comparison}. Our FedWCA outperforms all other methods n all source domains for Digit-Five and PACS, highlighting that merely extending existing FL methods to FFREEDA is insufficient and confirming our method's effectiveness. FedProx uses basic averaging for aggregation, FedAMP's parameter similarity weighting is unsuitable for our unlabeled adaptation, and IFCA's cluster model selection is unstable due to the loss calculation based on pseudo-labels.
}


\begin{algorithm*}[t]
  \caption{FedWCA (Federated learning with weighted cluster aggregation)}
  \label{alg:fedwca}
  \SetKwFunction{FMain}{}
  \SetKwProg{Fn}{ClientLocalAdaptation}{:}{}
  \Input{$K$ clients, $k$-th client's local unlabeled data $D_k=\{x_i\}_{i=1}^{N_k}$, $k$-th client's initial local model $h_k^{\rm init} = g_k^{\rm init} \circ f_k^{\rm init}$, source model $h_S=g_S \circ f_S$, local epochs $E$, total communication round $R$, temperature parameters $T_a$, $T_b$, balancing parameters $\lambda$, $\mu$}
  \Output{$K$ personalized models $h_k = g_S \circ f_{c_k}$}
  \For{$r=0, \dots R-1$}{%
    \eIf{$r=0$}
        {
            \textbf{Client:} \\
            \quad Initialize the local model: $h_k^{\rm init} = g_k^{\rm init} \circ f_k^{\rm init} \gets h_S = g_S \circ f_S$ \\
            \quad $f_k =$ \textbf{ClientLocalAdaptation}($r$, $D_k$, $f_k^{\rm init}$, $g_k^{\rm init}$, $E$, $\lambda$) \\
            \quad Send $f_k$ to the server \\
            \textbf{Server:} \\
            \quad Cluster $K$ clients into $C$ clusters based on the first layers' parameters of $f_k$ by using FINCH: $k \mapsto c_k$ \\
            \quad Create cluster models $f_c$ by averaging the local models $f_k$ in each cluster $c \in \{1,\dots,C\}$ \\
            \quad Initialize soft cluster models $\tilde{f}_c = f_c$
        }
        {
            \textbf{Client:} \\
            \quad Calculate $\bm{\alpha}_k$ for $\tilde{f}_c$ and $\bm{\beta}_k$ for $f_{c_k}$ and $\bar{f}_c = \sum_{c}\alpha_{k,c} \tilde{f}_c$ by \cref{cluster_weights} and \cref{snd_weight}\\
            \quad Set an initial model: $f_k^{\rm init}=\beta_{k,0} f_{c_k} + \beta_{k,1} \sum_{c}\alpha_{k,c} \tilde{f}_c$ \\
            \quad $f_k =$ \textbf{ClientLocalAdaptation}($r$, $D_k$, $f_k^{\rm init}$, $f_{c_k}$, $g_k^{\rm init}$, $E$, $T_a$, $T_b$, $\lambda$, $\mu$) \\
            \quad Send $f_k$, $\bm{\alpha}_k$, and $\bm{\beta}_k$ to the server \\
            \textbf{Server:} \\
            \quad Update cluster models $f_c$ by averaging the local models $f_k$ in each cluster $c \in \{1,\dots,C\}$ \\
            \quad Calculate $A_{c \rightarrow c'}$ and $B_{c,i}$ by averaging $\alpha_{k,c'}$ and $\beta_{k,i}$ across clients within each cluster $c$ \\
            \quad Update soft cluster models: $\tilde{f}_c = B_{c,0}f_c + B_{c,1}\sum_{c'\in\mathcal{C}}A_{c'\rightarrow c}f_{c'}$ \\
        }
        \textbf{Server:} Send $f_{c_k}$ and $\tilde{f}_c$ for every $c$ to the client $k$
  }
\end{algorithm*}

\begin{algorithm*}[t]
  \caption{ClientLocalAdaptation}
  \label{alg:local_adaptation}
  \Input{Current round $r$, unlabeled data $D=\{x_i\}_{i=1}^{N}$, initial model $f^{\rm init}$, cluster model $f_c$, classifier $g$, local epochs $E$, temperature parameters $T_a, T_b$, balancing parameters $\lambda, \mu$.}
  \Output{Trained model $f$}

  \eIf{$r=0$}
    {
        Compute class-wise prototypes $p_m$ for each class $m$ by $g \circ f^{\rm init}$ and assign each sample $x \in D$ pseudo-labels $\hat{y}$ via \cref{prototype}, and generate a pseudo-labeled dataset $\hat{D}=\{x_i,\hat{y}_i\}_{i=1}^N$
    }
    {
        Compute class-wise prototypes $p_m$ and $q_m$ for each class $m$ by $g \circ f^{\rm init}$ and $g \circ f_c$, and assign each sample $x \in D$ pseudo-label $\hat{y}_{\rm init}$ and $\hat{y}_c$ via \cref{prototype}, respectively
        
        Compute normalization factors $p$, $q$: $$p=\sum_{m \ne m'}\frac{\cos(p_m,p_{m'})}{M(M-1)}, \ \ q=\sum_{m \ne m'}\frac{\cos(q_m,q_{m'})}{M(M-1)}$$
        
        Assign each sample $x \in D$ pseudo-label $\hat{y}=\hat{y}_{\rm init}$ if $p_m/p \geq q_m/q$, and $\hat{y}=\hat{y}_c$ otherwise
        
        Generate pseudo-labeled datasets $\hat{D}^{mat}$ and $\hat{D}^{mis}$: $$\hat{D}^{mat}=\{(x,\hat{y})|x \in D \ s.t.\  \hat{y}_{\rm init}=\hat{y}_c)\}, \quad \hat{D}^{mis}=\{(x,\hat{y})|x \in D \ s.t.\  \hat{y}_{\rm init}\ne\hat{y}_c)\}$$
        
        Generate a mixed dataset $\hat{D}^{mix}$ and a pseudo-labeled dataset $\hat{D}=\hat{D}^{mat} \cup \hat{D}^{mix}$: $$\hat{D}^{mix}=\{((1-\mu)x+\mu x',\hat{y})|(x,\hat{y}) \in \hat{D}^{mis}, {\rm randomly\ sampled}\ (x',\hat{y}) \in \hat{D}^{mat}\ {\rm for\ each}\ (x,\hat{y})\}$$
    }
  
  Initialize a model: $f = f^{\rm init}$ \\
        \For{$e=0, \dots, E-1$}
        {
        Update $f$ with SGD by minimizing the loss function $\mathcal{L}(g \circ f) = \mathcal{L}_{\rm IM}(g \circ f;D) + \lambda \mathcal{L}_{\rm CE}(g \circ f;\hat{D})$ defined in \cref{loss_function} while fixing $g$
        }
        \textbf{return} $f$ 
\end{algorithm*}

\end{document}